\documentclass{article}

\PassOptionsToPackage{numbers, compress}{natbib}

 \usepackage[main, final]{neurips_2026}

\usepackage[utf8]{inputenc} 
\usepackage[T1]{fontenc}    
\usepackage{hyperref}       
\usepackage{url}            
\usepackage{booktabs}       
\usepackage{array}          
\usepackage{amsfonts}       
\usepackage{nicefrac}       
\usepackage{microtype}      
\usepackage[dvipsnames]{xcolor}  
\usepackage{amsmath}
\usepackage{graphicx}
\usepackage{enumitem}
\usepackage{subcaption}

\hypersetup{
  colorlinks=true,
  citecolor=RoyalBlue,
  linkcolor=BrickRed,
  urlcolor=blue!60!black,
}

\title{Jet-Long: Efficient Long-Context Extension with Dynamic Bifocal RoPE}

\author{%
  Haozhan Tang, Zerui Wang, Yuxian Gu, Song Han, Han Cai \\
  NVIDIA \\
  \texttt{hcai@nvidia.com} \\
  \url{https://github.com/jet-ai-projects/jet-long}
}

\begin{document}

\maketitle

\begin{abstract}
Modern LLMs are increasingly deployed in long-context applications such as retrieval-augmented generation, repository-level coding, and agentic workflows whose accumulated reasoning and tool traces routinely push the input an order of magnitude past the pretraining window, making zero-shot context extension the dominant deployment path for open-weight checkpoints.
The dominant zero-shot methods (YaRN, Self-Extend, DCA)~\citep{peng2023yarn,jin2024llm,an2024training} fix a single rescaling factor up front, so an aggressive factor sacrifices short-context fidelity while a conservative one breaks down at long contexts; recent length-aware variants~\citep{xu2025extending,zhang2025lampe} adapt the mapping, but with a fitted or distance-dependent schedule.
We propose \textbf{Jet-Long}, a tuning-free zero-shot method that pairs a local RoPE-faithful window with a long-range window whose rescaling factor adapts dynamically to the current sequence length via a parameter-free analytic schedule, recovering the base model exactly at short inputs while extrapolating cleanly at long ones.
An inclusion--exclusion attention merge and on-the-fly RoPE correction enable a fused CuTe implementation. On H100 at 64K--128K, prefill retains $83$--$88\%$ of FlashAttention-3 throughput across the evaluated Qwen3 sizes and $88$--$93\%$ of a matched CuTe control; Qwen3-8B single-batch generation reaches $1.04$--$1.08\times$ FlashAttention-3 throughput.
On Qwen3-1.7B/4B/8B~\citep{yang2025qwen3} up to 128K context, Jet-Long leads RULER by $+4.79$/$+2.18$/$+2.03$~pp over the strongest baseline at 1.7B/4B/8B, achieves the best overall accuracy on HELMET-RAG (a benchmark identified by HELMET as the most efficient predictor of downstream long-context performance~\citep{yen2024helmet}) and attains the lowest PG-19 perplexity.
Additional evaluations cover Meta-Llama-3-8B, post-trained Qwen3 checkpoints, and the hybrid Jet-Nemotron architecture~\citep{gu2025jet}, supporting broader applicability without retraining. The local-window hyperparameter remains robust across the tested settings.
\end{abstract}

\begin{figure*}[ht]
    \centering
    \includegraphics[width=\textwidth,trim=0 0 277bp 0,clip]{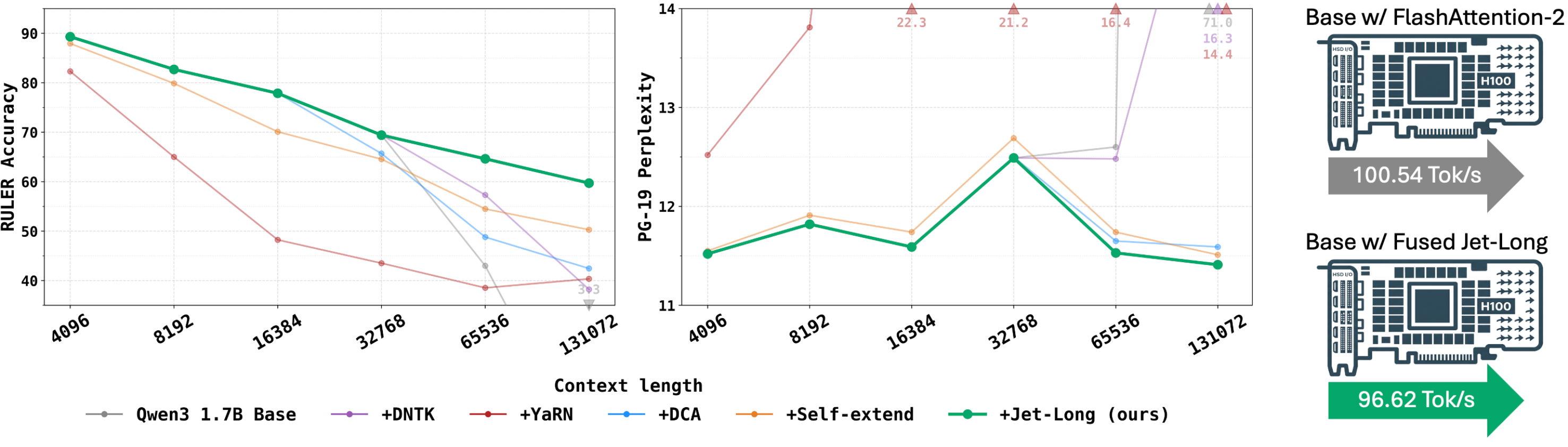}
    \caption{Qwen3-1.7B-Base RULER accuracy, averaged over all 13 tasks, and PG-19 perplexity at the six main context lengths. Jet-Long improves extended-context performance while preserving the base attention rule within the native window. The revised H100 efficiency comparison, with FA3 as the primary reference and a matched CuTe control, is reported in Section~\ref{sec:efficiency}.}
    \label{fig:intro}
\end{figure*}

\section{Introduction}
\label{sec:intro}
Large language models (LLMs) are now deployed in long-context applications including long-document QA, repository-level code understanding, retrieval-augmented generation, and multi-step agentic workflows~\citep{touvron2023llama,liu2025deepseek,liu2025comprehensive,liu2025thus,hsieh2024ruler,jimenez2023swe,gao2023retrieval,yu2024defense,wang2024survey}. The pressure is most severe in agentic LLMs that interleave reasoning, planning, and tool use across many turns~\citep{yao2022react,schick2023toolformer,liu2023agentbench}, and in coding agents operating over real software repositories~\citep{yang2024sweagent,hong2024metagpt}, where source code, execution traces, and tool outputs routinely accumulate to 100K+ tokens per task.

Training directly at long context remains expensive: efficient kernels like FlashAttention~\citep{dao2022flashattention} and Ring Attention~\citep{liu2023ring} make memory linear but compute stays quadratic in sequence length, and long-context data is scarce while long-context fine-tuning often degrades short-context behavior~\citep{an2024does,fu2024data}. Many models are therefore pretrained at a moderate window (4K--32K tokens) and expected to handle longer inputs at inference, a setting known as \emph{context extension}~\citep{press2021train,ding2024longrope}. \emph{Zero-shot context extension} (without fine-tuning) has become the dominant deployment mode for open-weight LLMs~\citep{yang2025qwen3,agarwal2025gpt,liu2024deepseek}, since a single released checkpoint must support arbitrary downstream context lengths.

Vanilla Transformer-based LLMs fail to generalize beyond the training window~\citep{press2021train,chen2023extending,liu2023scaling},
with two failure modes (out-of-distribution RoPE rotations and softmax-attention diffusion) detailed in Section~\ref{sec:related}.
A growing body of zero-shot methods (NTK~\citep{bloc972023ntk} / Dynamic NTK~\citep{emozilla2023dyntk}, YaRN~\citep{peng2023yarn}, Self-Extend~\citep{jin2024llm}, DCA~\citep{an2024training}) addresses one or the other; these constitute the zero-shot baselines we benchmark against.

We propose \textbf{Jet-Long}, a tuning-free zero-shot context extension method that pairs a local RoPE-faithful window with a long-range window whose rescaling factor is \emph{dynamic} in the current sequence length. Unlike the fixed configurations of YaRN, Self-Extend, and DCA, and unlike length-adaptive maps with fitted or distance-dependent schedules~\citep{xu2025extending,zhang2025lampe}, Jet-Long derives the factor analytically from the current sequence length and the pretrained window, with no per-model fitting. Cache correction and kernel fusion make the adaptive mapping practical at inference.

Our contributions are:
\begin{itemize}
    \item \textbf{Jet-Long}, a tuning-free bifocal context-extension method that sets the remote group size to the minimal integer compression $G=\max(1,\lceil L/w_{\text{pretrained}}\rceil)$ at the current sequence length $L$, keeping remote relative rotations on the pretrained grid without per-model fitting and mathematically recovering base attention within the native context.
    \item An inclusion--exclusion attention merge with on-the-fly RoPE correction that avoids rewriting cached keys. CuTe fusion reaches $0.83$--$0.88\times$ FlashAttention-3~\citep{shah2024flashattention3} prefill throughput on H100 at 64K--128K across three model sizes, and $0.88$--$0.93\times$ a matched CuTe control without bifocal attention; 8B decode reaches $1.04$--$1.08\times$ FA3. These measurements separate routing cost from improvements in the underlying attention implementation.
    \item Evaluation on Qwen3-1.7B/4B/8B-Base up to 128K: Jet-Long leads the strongest compared zero-shot baseline on RULER~\citep{hsieh2024ruler} by $+4.79/+2.18/+2.03$~pp, has the best aggregate HELMET-RAG~\citep{yen2024helmet} score, and the lowest PG-19~\citep{rae2019compressive} perplexity at each size. Additional three-seed evaluations test transfer to Meta-Llama-3-8B and post-trained Qwen3; Jet-Nemotron~\citep{gu2025jet} demonstrates applicability to hybrid attention.
\end{itemize}

\section{Related work}
\label{sec:related}

\subsection{Why RoPE-based LLMs fail to extrapolate}
\label{sec:related-fail}
Most modern open-weight LLMs use Rotary Position Embedding (RoPE)~\citep{su2024roformer}, which applies per-position rotations across geometrically spaced frequencies so attention depends only on relative position; earlier relative-position schemes such as ALiBi~\citep{press2021train}, T5's relative-position bias~\citep{raffel2020t5}, and iRPE~\citep{wu2021irpe} have largely been supplanted. Two failure modes prevent these models from extrapolating beyond their training window.

\paragraph{(i) Position out-of-distribution.} At sequence positions never seen during training, the low-frequency RoPE components produce rotation angles outside the training distribution, causing attention scores to behave erratically~\citep{chen2023extending,peng2023yarn,liu2023scaling}.

\paragraph{(ii) Attention diffusion and positional bias.} As the key set grows, the softmax distribution flattens, dispersing probability mass over irrelevant tokens~\citep{han2024lm,peng2023yarn}; separately, models exhibit a U-shaped positional attention bias that under-attends to middle-context information, degrading retrieval accuracy for centrally placed evidence~\citep{liu2024lost}.

These motivate two complementary zero-shot axes: interpolating RoPE frequencies or position indices to keep rotation angles in-distribution~\citep{chen2023extending,bloc972023ntk,emozilla2023dyntk,peng2023yarn,jin2024llm,an2024training,li2025training,su2023rerope}, and attention penalties or temperature scaling to counteract softmax diffusion~\citep{peng2023yarn,han2024lm,li2025training}. Jet-Long targets position-OOD via dynamic aliasing onto the pretrained rotation grid.

\subsection{Zero-shot context extension methods}
\label{sec:related-zero-shot}

\paragraph{Frequency-rescaling methods.}
Position Interpolation (PI)~\citep{chen2023extending} linearly compresses positions into the pretrained range. NTK-aware scaled RoPE~\citep{bloc972023ntk} increases the RoPE base to preserve high-frequency components while interpolating low-frequency ones; Dynamic NTK (DNTK)~\citep{emozilla2023dyntk} makes that base a function of the current sequence length so the scaling adapts at decode time. YaRN~\citep{peng2023yarn} combines per-dimension frequency partitioning with an attention-temperature correction. Beyond pure rescaling, ReRoPE~\citep{su2023rerope} caps relative distances past a window threshold (implemented as a two-pass within/beyond-window attention blend at prefill), and GALI~\citep{li2025training} interpolates at the attention-logit level rather than the embedding.

\paragraph{Grouped / chunked-position methods.}
A second line reuses only in-distribution position indices. Self-Extend~\citep{jin2024llm} pairs a neighbor window with a grouped window in which blocks of tokens share a single position index. Dual Chunk Attention (DCA)~\citep{an2024training} partitions the sequence into chunks and uses asymmetric query/key indices in the cross-chunk component so all relative distances stay within the pretrained range. LM-Infinite~\citep{han2024lm} earlier introduced a $\Lambda$-shaped mask retaining an attention sink and a recent window.

\paragraph{Length-adaptive remapping.}
Closest to our setting, several training-free methods also condition the position map on input length. AdaGroPE~\citep{xu2025extending} grows the reuse count of relative positions with distance; LaMPE~\citep{zhang2025lampe} applies a piecewise map over three sequence regions whose interior compression ratio is a per-model fitted sigmoid of the input length; SELF~\citep{dang2025self} grows Self-Extend's group size logistically. LaMPE's Appendix~E also reports additional latency and memory use relative to FlashAttention on A800, motivating attention to execution cost; this is a separate hardware setup from our matched H100 study.

Jet-Long sits in the grouped-position family, with the group size derived from the current sequence length rather than fixed or fitted: identity within the native window, and past it the minimal integer $G$ that keeps remote relative rotations on the pretrained grid (Section~\ref{sec:dynamic}). It therefore reduces mathematically to base attention for $L \le w_{\text{pretrained}}$, and its correction rotation (Section~\ref{sec:kv-cache}) preserves a base-position key cache when $G$ changes mid-generation, without recomputing historical hidden states.

\subsection{Long-context training, architectures, and benchmarks}
\label{sec:related-rest}

Training-based approaches extend the window via continued pretraining~\citep{ding2024longrope} but face quadratic FLOP costs (only partially mitigated by efficient kernels~\citep{dao2022flashattention,liu2023ring}) and scarce long-context data with short-context regression risk~\citep{an2024does,fu2024data}, motivating the zero-shot setting. Our evaluation does not include a matched comparison with training-based methods such as LongRoPE; comparisons reported for other backbones and protocols do not establish their performance relative to Jet-Long (Appendix~\ref{app:limitations}).

Alternative architectures replace dense softmax with sparse attention~\citep{beltagy2020longformer,zaheer2020bigbird}, linear or kernel-based variants~\citep{katharopoulos2020lineartransformer,choromanski2021performer}, or state-space models~\citep{gu2023mamba}; sparser or non-softmax distributions naturally curb the attention diffusion and lost-in-the-middle bias that plague dense softmax at long context. A complementary line removes positional encoding entirely: NoPE outperforms explicit position encodings on length-generalization benchmarks~\citep{kazemnejad2023nope}, and open-weight models such as Kimi K2~\citep{kimi2025k2} and NVIDIA Nemotron Nano 2~\citep{nvidia2025nemotron} adopt NoPE in their hybrid layers. Both lines typically require training from scratch or substantial fine-tuning and therefore complement rather than compete with our zero-shot setting; Jet-Long itself extends smoothly to hybrid designs (Section~\ref{sec:hybrid}, Jet-Nemotron~\citep{gu2025jet}).

We evaluate on RULER~\citep{hsieh2024ruler} (13 retrieval, tracking, extraction, and QA tasks), HELMET-RAG~\citep{yen2024helmet} (the HELMET study's best overall predictor of downstream long-context performance), and PG-19~\citep{rae2019compressive} long-form perplexity, against four zero-shot baselines. Additional evaluations cover Meta-Llama-3-8B and post-trained Qwen3 (Appendices~\ref{app:cross_family} and~\ref{app:posttrained}), and applicability to the hybrid Jet-Nemotron backbone (Section~\ref{sec:hybrid}).

\section{Methodology}
\label{sec:method}

Fixed rescaling and grouping factors force a tradeoff between short-context fidelity and long-context reach; recent length-adaptive maps~\citep{xu2025extending,zhang2025lampe} relax it, but with a fitted or distance-dependent schedule.
Jet-Long instead derives the factor analytically from the current sequence length (Section~\ref{sec:dynamic}) and couples the two-window computation with cache correction and fused attention (Sections~\ref{sec:kv-cache}--\ref{sec:prefill}). Section~\ref{sec:efficiency} measures the resulting inference cost against FlashAttention-3 and a matched CuTe control.

We build on the dual-window (bifocal) decomposition of Self-Extend~\citep{jin2024llm} (a related but architecturally three-way ancestor is DCA~\citep{an2024training}): a local window of size $w_0$ that retains classic RoPE, and a remote window governed by a rewritten position mapping $f(\cdot)$ that maps positions back into the training range. Grouping shares only positional coordinates: every token retains a separate key and value, with no token merging or removal. For a query at position $q$ and a key at position $k$, the pre-softmax attention score is
\begin{equation}
S(q, k) =
\begin{cases}
\text{RoPE}(\mathbf{x}_q, q)^\top \text{RoPE}(\mathbf{x}_k, k) & \text{if } q - k \le w_0, \\
\text{RoPE}(\mathbf{x}_q, f(q))^\top \text{RoPE}(\mathbf{x}_k, f(k)) & \text{if } q - k > w_0.
\end{cases}
\label{eq:two-window}
\end{equation}
Let $L$ denote the current sequence length and $w_{\text{pretrained}}$ the pretrained context window. Fresh prefill uses one $G$ determined by the full prompt length; decoding updates $L$ to include the cached prefix and current token. When $L \le w_{\text{pretrained}}$, $f(x) = x$ and Eq.~\eqref{eq:two-window} reduces to the unmodified base attention. Jet-Long's contribution lies in the choice of $f(\cdot)$ and its cache-aware, fused realization.

\begin{figure*}[t]
\centering
\begin{subfigure}[t]{0.38\textwidth}
    \centering
    \includegraphics[width=\textwidth]{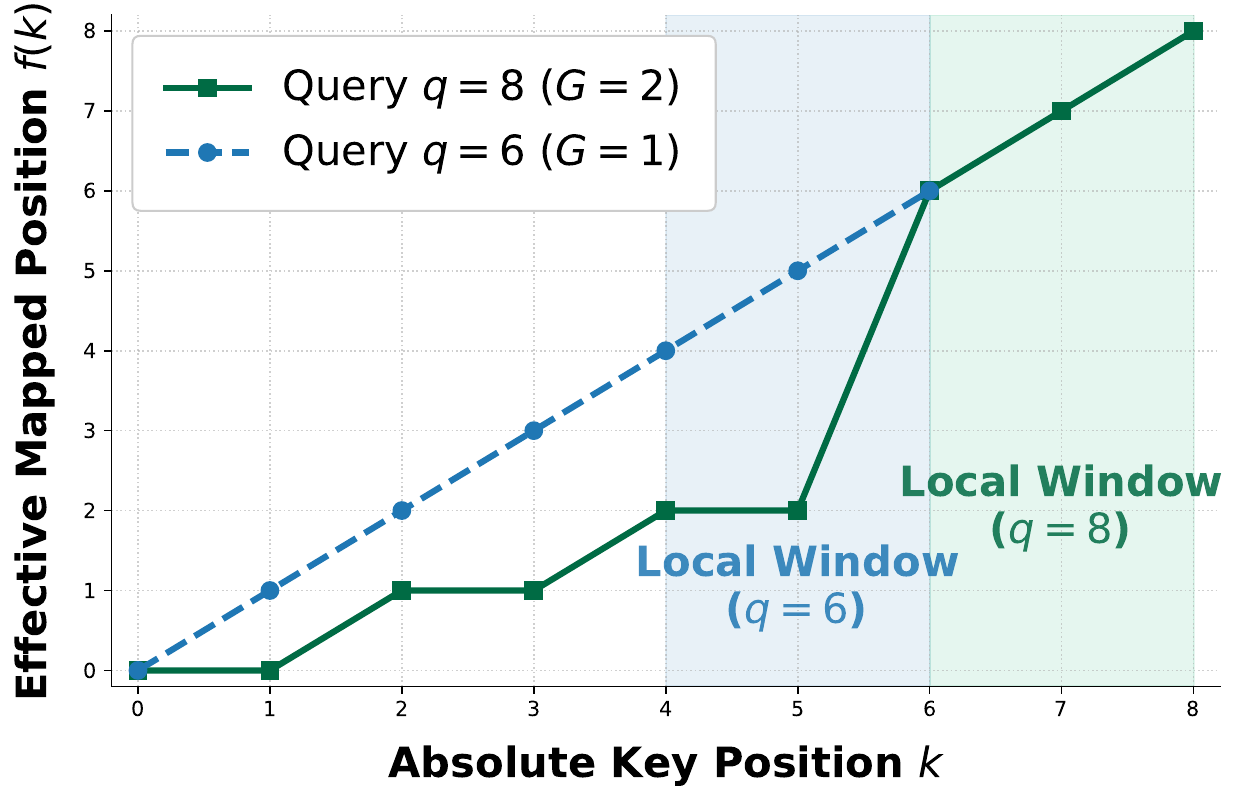}
    \caption{Dynamic bifocal mapping}
    \label{fig:method-a}
\end{subfigure}
\hfill
\begin{subfigure}[t]{0.30\textwidth}
    \centering
    \includegraphics[width=\textwidth]{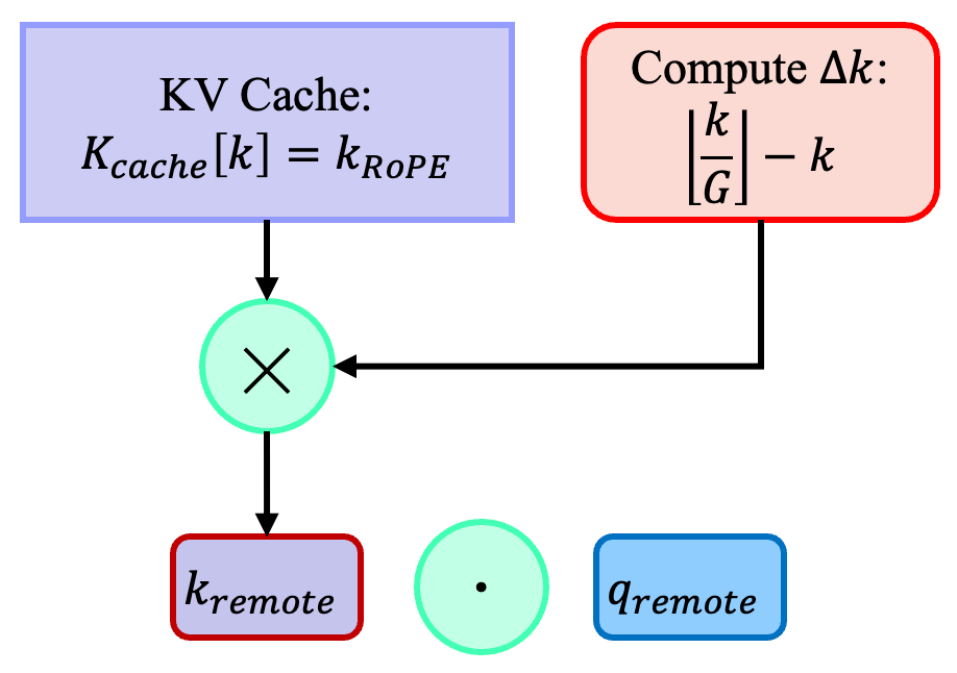}
    \caption{KV cache correction at decode}
    \label{fig:method-b}
\end{subfigure}
\hfill
\begin{subfigure}[t]{0.30\textwidth}
    \centering
    \includegraphics[width=\textwidth]{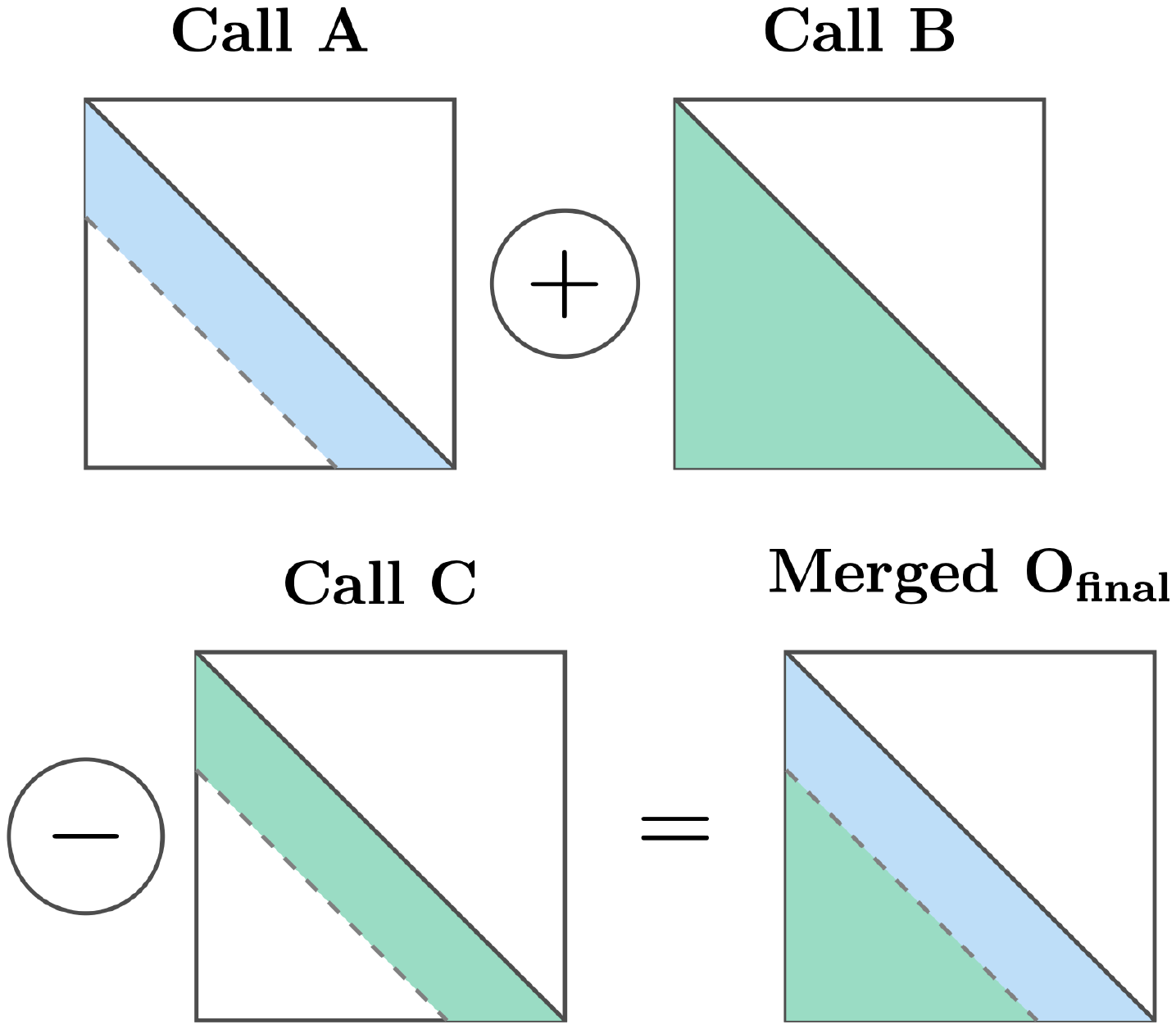}
    \caption{Inclusion--exclusion prefill}
    \label{fig:method-c}
\end{subfigure}
\caption{Overview of Jet-Long. \textbf{(a)} The local window of width $w_0$ keeps classic RoPE; remote keys use $f(x) = \lfloor x / G \rfloor$ with $G = \max(1, \lceil L / w_{\text{pretrained}} \rceil)$, adapting to context length $L$. \textbf{(b)} Cached keys retain original RoPE coordinates. Correction rotations ($\Delta q$ on the active query, $\Delta k$ on each cached key) recover remote phases without rewriting stored keys or recomputing historical hidden states. \textbf{(c)} Prefill combines three attention computations (full remote, local-only-with-RoPE, local-only-with-remap) by inclusion--exclusion, using LogSumExp statistics; CuTe fusion reduces execution overhead.}
\label{fig:method}
\end{figure*}

\subsection{Dynamic extrapolation factor}
\label{sec:dynamic}

To keep remote-window relative rotation angles on the pretrained position grid, the group size $G$ must scale with the current sequence length $L$, as illustrated in Figure~\ref{fig:method-a}.

The natural dynamic factor is, for continuous-style RoPE rescaling (e.g., DNTK~\citep{emozilla2023dyntk} and dynamic-YaRN~\citep{peng2023yarn} variants), the scaling ratio $s = \max(1, L/w_{\text{pretrained}})$; for discrete grouped methods (e.g., Self-Extend~\citep{jin2024llm}), it is an integer group size $G$. Jet-Long uses discrete grouping because pretraining uses a finite, discrete set of relative RoPE rotation angles; aliasing positions onto that grid retains integer relative positions within the training range. This is a statement about positional phases, not a guarantee that long-context hidden states follow their training distribution. We ablate this choice against continuous frequency interpolation in Section~\ref{sec:alias_vs_freq}. Whereas DNTK~\citep{emozilla2023dyntk} adapts the RoPE base $\beta$ (per-frequency), Jet-Long adapts the discrete group size $G$ on the position-aliasing axis.

To maximize positional resolution, $G$ is the smallest integer that keeps the compressed sequence within the pretrained window $w_{\text{pretrained}}$:
\begin{equation}
G = \max \left( 1, \left\lceil \frac{L}{w_{\text{pretrained}}} \right\rceil \right)
\label{eq:group-size}
\end{equation}

The remote mapping is a floor division of absolute positions:
\begin{equation}
f(x) = \left\lfloor \frac{x}{G} \right\rfloor
\end{equation}

By recomputing $G$ as the sequence grows, Jet-Long applies the minimum integer compression that keeps every remote position within the pretrained range, maximizing positional resolution within this mapping family. LaMPE~\citep{zhang2025lampe} likewise applies floor-based remapping in its middle region, but derives the middle-region compression ratio from an empirically fitted sigmoid of the input length, with model-specific fits presented for Llama2 and Llama3, whereas Eq.~\eqref{eq:group-size} is determined analytically by $L$ and $w_{\text{pretrained}}$ with no fitted parameters.

\subsection{Key-value cache management and correction rotation}
\label{sec:kv-cache}

Dynamic extrapolation poses a cache-management challenge: cached keys must be interpreted under the current $G$ even when it changes during generation.

As depicted in Figure~\ref{fig:method-b}, Jet-Long avoids this overhead by maintaining a strict invariant: the cache stores only uncompressed base keys at their exact absolute positions $k$,
\begin{equation}
\mathbf{k}_{\text{cache}}[k] = \text{RoPE}(\mathbf{x}_k, k)
\end{equation}

When the remote window needs the query at $f(q)$ and the key at $f(k)$, instead of recomputing those vectors from scratch we apply a correction rotation on the fly using the position offsets
\begin{equation}
\Delta q = f(q) - q = \left\lfloor \frac{q}{G} \right\rfloor - q, \qquad
\Delta k = f(k) - k = \left\lfloor \frac{k}{G} \right\rfloor - k.
\end{equation}
This relies on standard RoPE composing additively in angle, $R_a R_b = R_{a+b}$ (per-position scalings beyond a pure rotation would break this and require recomputing keys from $\mathbf{x}_k$). Applying $\text{RoPE}(\cdot, \Delta)$ to a vector encoding position $p$ therefore produces the vector at position $p+\Delta$, so we apply the offset directly to the active query and cached keys before attention:
\begin{equation}
\mathbf{q}_{\text{remote}} = \text{RoPE}(\mathbf{q}_{\text{local}}, \Delta q), \qquad
\mathbf{k}_{\text{remote}} = \text{RoPE}(\mathbf{k}_{\text{cache}}[k], \Delta k).
\end{equation}

The correction reconstructs transient remote views without rewriting the stored cache. It exactly recovers the current RoPE phases for the cached key vectors, but does not recompute historical hidden states or values. Retained-cache decoding is therefore not equivalent to a full re-prefill of the prefix under the new $G$. Appendix~\ref{app:cache_validation} measures this approximation through a direct re-prefill comparison and a stratified analysis of group-size transitions.

\subsection{Inclusion--exclusion prefill}
\label{sec:prefill}

To avoid materializing the full quadratic attention matrix during long-sequence prefill, Jet-Long combines inclusion--exclusion with FlashAttention's~\citep{dao2022flashattention} LogSumExp statistics. The merge uses three standard attention calls (Figure~\ref{fig:method-c}), each returning an output $\mathbf{O}_X$ and an LSE vector $\boldsymbol{\ell}_X$: \textbf{(A)}~sliding-window attention (local $w_0$) with base queries and keys, \textbf{(B)}~full causal attention with remote queries and keys, and \textbf{(C)}~sliding-window attention (local $w_0$) with remote queries and keys. Calls~B and~C apply the same remote rotation to local keys, so their local-subset contributions cancel in exact arithmetic: $W_B \mathbf{O}_B - W_C \mathbf{O}_C$ (resp. $W_B - W_C$) collects only the remote-only term, while Call~A supplies the base-local term. Stabilizing element-wise (per query position) via $M = \max(\boldsymbol{\ell}_A, \boldsymbol{\ell}_B, \boldsymbol{\ell}_C)$ and weights $W_X = \exp(\boldsymbol{\ell}_X - M)$, the final output (numerator and denominator accumulated in FP32) is
\begin{equation}
\mathbf{O}_{\text{final}} = \frac{W_A \mathbf{O}_A + W_B \mathbf{O}_B - W_C \mathbf{O}_C}{W_A + W_B - W_C}.
\label{eq:inclusion-exclusion}
\end{equation}
This algebraically exact merge implements one joint softmax over the local and remote windows without a dense boolean mask. It retains dense attention and its quadratic arithmetic cost; fusion reduces implementation overhead. Subtractive attention merges can suffer cancellation~\citep{yao2026focus}. Here subtraction occurs in linear space, the shift by $M$ bounds weights in $(0,1]$ in exact arithmetic, and FP32 accumulation reduces roundoff, without providing an unconditional numerical-stability guarantee.

\section{Experiments}
\label{sec:exp}

\subsection{Setup}
\label{sec:setup}
\paragraph{Models.}
Our primary evaluation is on Qwen3-1.7B/4B/8B-Base~\citep{yang2025qwen3}, each with a 32{,}768-token (32K) native window, extended to 131{,}072 (128K) without fine-tuning. We also evaluate Meta-Llama-3-8B-Base~\citep{meta2024llama3}, post-trained Qwen3-1.7B/8B with their chat templates, and the hybrid Jet-Nemotron-2B/4B models~\citep{gu2025jet}, which interleave softmax and linear-attention layers. The main Qwen3 comparisons use $w_0=2048$; baseline configurations and additional sensitivity tests are in Appendices~\ref{app:baseline_hp} and~\ref{app:baseline_sensitivity}.

\paragraph{Long-context benchmarks.}
We report results on three complementary suites: (i) RULER~\citep{hsieh2024ruler}, a synthetic recall stress test averaging over thirteen tasks; (ii) HELMET-RAG~\citep{yen2024helmet}, averaged over the four default sub-tasks (NaturalQuestions, TriviaQA, PopQA, HotpotQA), which the HELMET study reports as the most efficient predictor of downstream long-context performance and which serves as our application-grounded benchmark; and (iii) PG-19~\citep{rae2019compressive}, on which we report long-form perplexity. RULER and HELMET-RAG accuracies are percentages, with gaps between methods reported in percentage points (pp); PG-19 is reported as token-level perplexity (ppl, lower is better). All RULER and HELMET-RAG generations use greedy decoding with the default RULER/HELMET prompts and per-task max-output-token limits; PG-19 is teacher-forced perplexity at stride 1024 over 100 books. Formal definitions and the geometric-mean aggregation used in the Avg columns are given in Appendix~\ref{app:metrics}. We further ablate the only Jet-Long hyperparameter, the local protected window size $w_0$, in Section~\ref{sec:w0_ablation}.

\paragraph{Inference efficiency.}
We benchmark fused CuTe prefill and single-batch generation on H100, using FlashAttention-3~\citep{shah2024flashattention3} as the primary hardware-appropriate reference, FA2~\citep{dao2023flashattention2} as a secondary reference, and FA4~\citep{zadouri2026flashattention4} as a complementary implementation. A matched CuTe control without bifocal attention separates the extension cost from the underlying kernel implementation.

\paragraph{Lengths and uncertainty.}
The main Base-model comparisons use six lengths: 4K, 8K, 16K, 32K, 64K, and 128K. Their additional 96K results are reported in the appendix. RULER uses 500 examples per task, except for the 256K--1M evaluation, which uses 100 (Section~\ref{sec:extreme}); the cross-family and post-trained evaluations use three independently generated datasets (seeds 42/44/45), reporting mean $\pm$ half-range. This measures dataset-sampling variation, not a confidence interval or training variance (Appendix~\ref{app:metrics}). Compute details and the frozen efficiency stack are in Appendices~\ref{app:compute} and~\ref{app:efficiency}.
\subsection{Main results on long-context extension}
\label{sec:main}

The average scores on the three benchmarks are reported in Table~\ref{tab:ruler_helmet}.
Jet-Long is best on every RULER and PG-19 column across all three Qwen3 sizes (over Base and the four extrapolation baselines), and best on HELMET-RAG at 4B and 8B (0.73 pp behind Self-Extend at 1.7B).
The RULER lead over the strongest baseline configuration in this table (DNTK at 1.7B; Self-Extend at 4B and 8B) is 4.79, 2.18, and 2.03 pp. Additional fixed-configuration sensitivity tests improve Self-Extend but preserve Jet-Long's lead at extended lengths (Appendix~\ref{app:baseline_sensitivity}).

\begin{table}[t]
\centering
\caption{Long-context performance on RULER (accuracy averaged over 13 tasks and six lengths: 4K, 8K, 16K, 32K, 64K, 128K), HELMET-RAG, and PG-19 perplexity (geometric mean over the same six lengths, lower is better), on three Qwen3 base sizes. Each Avg column averages the three model-size scores. Best per column in \textbf{bold}.}
\label{tab:ruler_helmet}
\renewcommand{\arraystretch}{0.9}
\setlength{\tabcolsep}{2.5pt}
\footnotesize
\begin{tabular*}{\textwidth}{@{\extracolsep{\fill}} l ccc c ccc c ccc c @{}}
\toprule
& \multicolumn{4}{c}{\textbf{RULER}} & \multicolumn{4}{c}{\textbf{HELMET-RAG}} & \multicolumn{4}{c}{\textbf{PG-19 ppl} ($\downarrow$)} \\
\cmidrule(lr){2-5} \cmidrule(lr){6-9} \cmidrule(lr){10-13}
\textbf{Method} & 1.7B & 4B & 8B & Avg & 1.7B & 4B & 8B & Avg & 1.7B & 4B & 8B & Avg \\
\midrule
Base            & 60.93 & 69.94 & 73.13 & \multicolumn{1}{c}{68.00} & 36.20 & 44.33 & 47.24 & \multicolumn{1}{c}{42.59} & 16.13 & 14.84 & 12.80 & \multicolumn{1}{c}{14.59} \\
DNTK~\citep{emozilla2023dyntk} & 69.14 & 79.75 & 83.54 & \multicolumn{1}{c}{77.48} & 41.27 & 50.81 & 55.55 & \multicolumn{1}{c}{49.21} & 12.60 & 10.78 & 9.13 & \multicolumn{1}{c}{10.84} \\
YaRN~\citep{peng2023yarn} & 52.99 & 70.11 & 78.49 & \multicolumn{1}{c}{67.20} & 32.24 & 43.63 & 53.41 & \multicolumn{1}{c}{43.09} & 16.39 & 12.08 & 9.91 & \multicolumn{1}{c}{12.79} \\
DCA~\citep{an2024training} & 67.80 & 80.19 & 81.08 & \multicolumn{1}{c}{76.36} & 41.77 & 51.91 & 56.12 & \multicolumn{1}{c}{49.93} & 11.77 & 9.89 & 8.77 & \multicolumn{1}{c}{10.14} \\
Self-Extend~\citep{jin2024llm} & 67.86 & 80.84 & 84.71 & \multicolumn{1}{c}{77.80} & \textbf{43.01} & 52.98 & 56.86 & \multicolumn{1}{c}{50.95} & 11.85 & 9.95 & 8.81 & \multicolumn{1}{c}{10.20} \\
Jet-Long (ours) & \textbf{73.93} & \textbf{83.02} & \textbf{86.74} & \multicolumn{1}{c}{\textbf{81.23}} & 42.28 & \textbf{53.61} & \textbf{57.34} & \multicolumn{1}{c}{\textbf{51.08}} & \textbf{11.72} & \textbf{9.85} & \textbf{8.73} & \multicolumn{1}{c}{\textbf{10.10}} \\
\bottomrule
\end{tabular*}
\end{table}

\begin{figure*}[ht]
    \centering
    \includegraphics[width=0.9\textwidth]{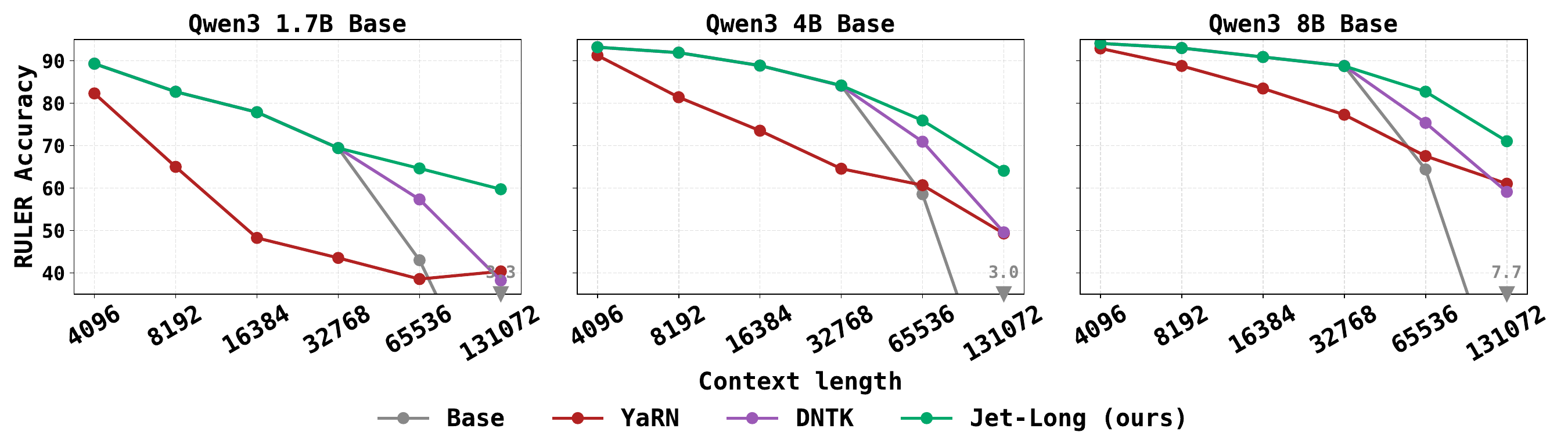}
    \caption{RULER accuracy as a function of context length, averaged over the 13 RULER tasks, on Qwen3-1.7B/4B/8B-Base. Jet-Long preserves the pretrained model's performance within the training window (32K) and outperforms YaRN and DNTK at all extended context lengths (the full comparison vs.\ DCA and Self-Extend is in Table~\ref{tab:ruler_helmet}).}
    \label{fig:ruler_by_length}
\end{figure*}

As Figure~\ref{fig:ruler_by_length} shows, Jet-Long matches or substantially outperforms YaRN and DNTK at every length across all three sizes for RULER accuracy by length. 
Table~\ref{tab:ppl_by_length} confirms the same pattern on PG-19: within the 32K training window, Jet-Long is mathematically equivalent to the base model (the dynamic factor reduces to identity, Section~\ref{sec:dynamic}); past 32K the bare base collapses (ppl of 71.00 / 104.66 / 79.37 at 128K for 1.7B / 4B / 8B) while Jet-Long stays at 11.41 / 9.62 / 8.51, the lowest among all extrapolation methods.

The per-task diagnostics in Appendix~\ref{app:ruler_tasks} show the largest gains on multi-key retrieval and variable tracking. Jet-Long does not lead every task: common-word extraction remains difficult for the 1.7B model. Full PG-19 results, including the additional 96K evaluation, are in Appendix~\ref{app:ppl_by_length}.

\subsection{Cross-family and post-trained evaluation}
\label{sec:generalization}

Table~\ref{tab:generalization} reports three-seed RULER results at 64K, 96K, and 128K. Meta-Llama-3-8B-Base has an 8K native window and RoPE base $\theta=500{,}000$, testing transfer beyond Qwen3. Jet-Long trails Self-Extend by 0.71~pp at 64K, within the observed sampling spread, and leads by 4.50/11.99~pp at 96K/128K. On post-trained Qwen3-1.7B, Jet-Long leads at all three displayed lengths. On post-trained Qwen3-8B, DNTK leads at 64K, while Jet-Long has the highest mean at 96K and 128K. These prompted-generation results broaden checkpoint coverage; they do not replace direct instruction-following or agent evaluations. Full results from 4K through 128K and additional baseline configurations are in Appendices~\ref{app:cross_family} and~\ref{app:posttrained}.

\begin{table}[t]
\centering
\caption{RULER accuracy: mean $\pm$ half-range over seeds 42/44/45, with 500 examples per task for all 13 tasks. Best mean per model and length in \textbf{bold}. $^\dagger$ marks at least 10\% empty generations in a seed; these scores include generation failures. Full seven-length results and baseline configurations are in Appendices~\ref{app:cross_family} and~\ref{app:posttrained}.}
\label{tab:generalization}
\footnotesize
\setlength{\tabcolsep}{2pt}
\renewcommand{\arraystretch}{1.0}
\begin{tabular*}{\textwidth}{@{\extracolsep{\fill}} l cccccc @{}}
\toprule
\textbf{Length} & \textbf{Unextended} & \textbf{DNTK} & \textbf{YaRN} & \textbf{DCA} & \textbf{Self-Extend} & \textbf{Jet-Long} \\
\midrule
\multicolumn{7}{c}{\textbf{Meta-Llama-3-8B-Base}} \\
64K & $0.00\pm0.00$ & $28.94\pm0.64$ & $31.87\pm0.37$ & $30.30\pm0.40$ & $\mathbf{58.28}\pm0.97$ & $57.57\pm1.17$ \\
96K & $0.00\pm0.00$ & $19.32\pm0.11$ & $26.15\pm0.89$ & $16.32\pm0.63$ & $47.86\pm1.38$ & $\mathbf{52.36}\pm1.64$ \\
128K & $0.00\pm0.00$ & $11.59\pm0.28$ & $12.21\pm0.79$ & $12.29\pm0.25$ & $31.86\pm0.99$ & $\mathbf{43.85}\pm0.67$ \\
\midrule
\multicolumn{7}{c}{\textbf{Qwen3-1.7B, post-trained}} \\
64K & $42.89\pm0.31$ & $51.78\pm0.35$ & $27.26\pm0.38^{\dagger}$ & $44.13\pm0.44$ & $48.51\pm1.76^{\dagger}$ & $\mathbf{54.88}\pm1.23$ \\
96K & $13.15\pm0.42$ & $36.28\pm0.32^{\dagger}$ & $34.36\pm0.14^{\dagger}$ & $41.07\pm0.65$ & $48.30\pm0.49$ & $\mathbf{50.50}\pm0.89$ \\
128K & $2.37\pm1.11$ & $27.70\pm1.04^{\dagger}$ & $25.78\pm0.19^{\dagger}$ & $38.51\pm1.03$ & $42.75\pm0.87^{\dagger}$ & $\mathbf{47.99}\pm0.90$ \\
\midrule
\multicolumn{7}{c}{\textbf{Qwen3-8B, post-trained}} \\
64K & $61.22\pm0.41$ & $\mathbf{77.63}\pm0.73$ & $64.85\pm0.46$ & $64.57\pm0.43$ & $73.65\pm0.79$ & $74.91\pm0.23$ \\
96K & $19.70\pm0.59$ & $70.56\pm0.70$ & $61.54\pm0.77$ & $62.78\pm0.23$ & $71.81\pm0.88$ & $\mathbf{72.94}\pm0.36$ \\
128K & $3.00\pm0.20$ & $63.62\pm0.64$ & $56.92\pm0.68$ & $61.05\pm0.46$ & $68.54\pm0.16$ & $\mathbf{68.88}\pm0.21$ \\
\bottomrule
\end{tabular*}
\end{table}

\subsection{Extension from 256K to 1M}
\label{sec:extreme}

We extend Qwen3-30B-A3B-Instruct-2507~\citep{qwen2025instruct2507} from its native 256K context to 512K and 1M. Table~\ref{tab:extreme_ruler} reports mean $\pm$ half-range over three independently generated dataset seeds, with 100 examples for each of the 13 RULER tasks per seed and length. Jet-Long achieves the highest mean at both extended lengths, exceeding DNTK by 2.16~pp at 512K and YaRN by 6.32~pp at 1M. The unextended model retains 67.22 at 512K and falls to 26.82 at 1M. Self-Extend is omitted because its two rotated key-cache views incur prohibitive memory or runtime costs in the standard implementation; a practical accuracy comparison requires a fused implementation.

\begin{table}[t]
\centering
\caption{RULER on Qwen3-30B-A3B-Instruct-2507: mean $\pm$ half-range over three dataset seeds, with 100 examples per task. The native context is 256K; 512K and 1M are $2\times$ and $4\times$ extensions. Best mean per length in \textbf{bold}.}
\label{tab:extreme_ruler}
\footnotesize
\setlength{\tabcolsep}{10pt}
\begin{tabular}{lccc}
\toprule
\textbf{Method} & \textbf{256K (native)} & \textbf{512K ($2\times$)} & \textbf{1M ($4\times$)} \\
\midrule
Unextended & $74.71\pm0.04$ & $67.22\pm0.32$ & $26.82\pm0.11$ \\
YaRN & $62.20\pm1.27$ & $59.84\pm0.75$ & $57.71\pm0.59$ \\
DNTK & $74.91\pm0.31$ & $69.83\pm0.27$ & $52.99\pm0.30$ \\
DCA & $\mathbf{75.36}\pm0.92$ & $65.70\pm0.14$ & $56.90\pm0.66$ \\
Jet-Long & $74.72\pm0.35$ & $\mathbf{71.99}\pm0.56$ & $\mathbf{64.03}\pm0.97$ \\
\bottomrule
\end{tabular}
\end{table}

\subsection{Hybrid attention extension results}
\label{sec:hybrid}

To test generalization beyond softmax-only transformers, we apply Jet-Long to the hybrid Jet-Nemotron architecture~\citep{gu2025jet}, which interleaves softmax and linear-attention layers. Table~\ref{tab:ruler_jetlm} reports per-length RULER accuracy on Jet-Nemotron-2B and 4B.

\providecommand{\imp}[1]{\textcolor{green!60!black}{(+#1 pp)}}
\begin{table}[t]
\centering
\caption{RULER accuracy over the six main context lengths for Jet-Long applied to the hybrid Jet-Nemotron architecture~\citep{gu2025jet}, against the bare base model. The two configurations are mathematically identical within the 32K native window; the green tag reports the absolute accuracy-point gap of Jet-Long over Base for $L > 32$K.}
\label{tab:ruler_jetlm}
\setlength{\tabcolsep}{2.5pt}
\renewcommand{\arraystretch}{1.05}
\scriptsize
\begin{tabular}{l cccccc c}
\toprule
\textbf{Method} & 4K & 8K & 16K & 32K & 64K & 128K & \emph{Avg} \\
\midrule
\textbf{Jet-Nemotron-2B} & 82.35 & 68.93 & 61.21 & 46.59 & 20.34 &  8.54 & \multicolumn{1}{c}{47.99} \\
+ Jet-Long & 82.35 & 68.93 & 61.21 & 46.58 & 40.33~\imp{19.99} & 33.78~\imp{25.24} & \multicolumn{1}{c}{55.53~\imp{7.54}} \\
\midrule
\textbf{Jet-Nemotron-4B} & 84.08 & 69.24 & 62.35 & 48.87 & 17.48 &  5.65 & \multicolumn{1}{c}{47.95} \\
+ Jet-Long & 84.08 & 69.24 & 62.35 & 48.88 & 39.93~\imp{22.45} & 33.14~\imp{27.49} & \multicolumn{1}{c}{56.27~\imp{8.33}} \\
\bottomrule
\end{tabular}
\end{table}

Within the native window, the construction reduces to base attention. At 128K, Jet-Long improves the hybrid models from 8.54 to 33.78 at 2B and from 5.65 to 33.14 at 4B ($+25.24/+27.49$ pp). The six-length averages improve by $7.54/8.33$ pp; Appendix~\ref{app:hybrid_full} includes 96K. This demonstrates architectural applicability to hybrid LLMs; the experiment compares against the unextended models and does not establish superiority over other hybrid extension methods.

\subsection{Ablation: local window size \texorpdfstring{$w_0$}{w0}}
\label{sec:w0_ablation}

Table~\ref{tab:w0_ablation} sweeps the protected local window on Qwen3-4B/8B at 64K and 128K; Appendix~\ref{app:window_full} adds 96K. Setting $w_0=0$ leaves self-attention locally and grouped remote attention active, reducing RULER to 1.08--4.91 across the full sweep. Local protection is therefore essential in this construction. Every $w_0\in\{512,1024,2048,4096\}$ stays within 2~pp of the best at each tested length and within 1~pp on the three-length average, supporting the default $w_0=2048$. The tested components serve distinct roles: local protection and remote mapping affect accuracy, while the fusion comparison in Section~\ref{sec:efficiency} isolates execution cost.

\providecommand{\worse}[1]{\textcolor{red!60!black}{(#1 pp)}}
\begin{table}[t]
\centering
\caption{Local-window ablation on Qwen3-4B/8B-Base at 64K and 128K. Cells report RULER accuracy; \textcolor{red!60!black}{red} values give the gap in pp from the row's best score (\textbf{bold}). Appendix~\ref{app:window_full} adds 96K and the three-length averages.}
\label{tab:w0_ablation}
\setlength{\tabcolsep}{1.2pt}
\renewcommand{\arraystretch}{0.9}
\scriptsize
\begin{tabular*}{\textwidth}{@{\extracolsep{\fill}} l ccccccc @{}}
\toprule
\textbf{Length} & $w_0{=}0$ & 256 & 512 & 1024 & 2048 & 4096 & 8192 \\
\midrule
\multicolumn{8}{l}{\textbf{Qwen3-4B-Base + Jet-Long}} \\
64K  & 4.10~\worse{-72.55} & \textbf{76.65} & 76.63~\worse{-0.02} & 76.49~\worse{-0.16} & 76.57~\worse{-0.09} & 75.91~\worse{-0.74} & 74.69~\worse{-1.96} \\
128K & 1.08~\worse{-64.55} & 64.13~\worse{-1.50} & \textbf{65.63} & 65.57~\worse{-0.05} & 64.74~\worse{-0.89} & 64.06~\worse{-1.56} & 63.52~\worse{-2.11} \\
\midrule
\multicolumn{8}{l}{\textbf{Qwen3-8B-Base + Jet-Long}} \\
64K  & 4.91~\worse{-78.60} & \textbf{83.51} & 83.23~\worse{-0.28} & 83.31~\worse{-0.20} & 83.24~\worse{-0.27} & 82.71~\worse{-0.80} & 81.40~\worse{-2.11} \\
128K & 1.55~\worse{-69.47} & 69.48~\worse{-1.55} & 70.49~\worse{-0.53} & 70.68~\worse{-0.35} & 71.01~\worse{-0.01} & \textbf{71.03} & 69.65~\worse{-1.38} \\
\bottomrule
\end{tabular*}
\end{table}

\subsection{Ablation: interpolate frequency or alias position}
\label{sec:alias_vs_freq}

Table~\ref{tab:alias_vs_freq} compares discrete position aliasing with YaRN-style frequency interpolation within the same bifocal construction on Qwen3-1.7B-Base. Aliasing leads by 6.99~pp at 64K and 4.30~pp at 128K. Our hypothesis is that aliasing's loss of within-group positional resolution explains some exceptions: tokens sharing a coordinate remain distinct content vectors, but their fine offsets cannot be distinguished by remote RoPE alone. At 128K, interpolation leads on MK-NIAH-2, MK-NIAH-3, CWE, and QA-2. This pattern motivates distance-adaptive combinations of interpolation and aliasing, but does not establish a causal explanation or validate a hybrid mapping. The mapping ablation itself is limited to 128K.

\begin{table}[t]
\centering
\caption{Ablation on the remote-window mapping for Qwen3-1.7B-Base: position aliasing (default Jet-Long) vs. YaRN-style frequency interpolation, reported per task at $L{=}65536$ and $L{=}131072$. Best per column within each length block in \textbf{bold} (ties shared); aliasing wins on the overall average at both lengths.}
\label{tab:alias_vs_freq}
\setlength{\tabcolsep}{3pt}
\renewcommand{\arraystretch}{1.0}
\scriptsize
\begin{tabular}{l rrrrrrrrrrrrr r}
\toprule
\textbf{Variant} & S1 & S2 & S3 & MK1 & MK2 & MK3 & MV & MQ & VT & CWE & FWE & QA1 & QA2 & \textbf{Avg} \\
\midrule
\multicolumn{15}{l}{\textbf{$L{=}65536$}} \\
Aliasing  & \textbf{100.00} & \textbf{100.00} & \textbf{100.00} & \textbf{93.20} & \textbf{61.80} & \textbf{1.60} & \textbf{84.45} & \textbf{82.25} & \textbf{73.56} & 0.32 & \textbf{75.47} & \textbf{39.40} & \textbf{28.00} & \multicolumn{1}{c}{\textbf{64.62}} \\
Frequency & \textbf{100.00} & 95.40 & 98.40 & 75.20 & 49.00 & 1.20 & 81.40 & 81.10 & 64.36 & \textbf{0.78} & 55.13 & 28.20 & 19.00 & \multicolumn{1}{c}{57.63} \\
\midrule
\multicolumn{15}{l}{\textbf{$L{=}131072$}} \\
Aliasing  & \textbf{100.00} & \textbf{98.60} & \textbf{100.00} & \textbf{91.40} & 24.00 & 0.20 & \textbf{75.90} & \textbf{76.30} & \textbf{70.12} & 0.30 & \textbf{82.40} & \textbf{33.40} & 23.60 & \multicolumn{1}{c}{\textbf{59.71}} \\
Frequency & \textbf{100.00} & 84.40 & 97.80 & 72.60 & \textbf{33.00} & \textbf{0.40} & 69.80 & 66.50 & 56.32 & \textbf{0.52} & 79.93 & 32.80 & \textbf{26.20} & \multicolumn{1}{c}{55.41} \\
\bottomrule
\end{tabular}
\end{table}

\subsection{Inference efficiency}
\label{sec:efficiency}

We measure end-to-end throughput on an H100 80GB with CUDA graphs, using a frozen software stack. We use Hopper-optimized FA3~\citep{shah2024flashattention3} as the primary H100 reference, FA2~\citep{dao2023flashattention2} as a secondary reference, and FA4~\citep{zadouri2026flashattention4} as a complementary prefill comparison. The matched control, labeled ``CuTe (without bifocal attention),'' omits Jet-Long's local/remote attention rule for context extension. Comparing it with fused Jet-Long measures the cost of that rule on the same backend. At extended lengths, fused and unfused Jet-Long retain the same bifocal attention rule, so their comparison measures the speedup from fusion. Table~\ref{tab:efficiency} reports representative Qwen3-8B-Base measurements; Appendix~\ref{app:efficiency} includes 96K results and the software configuration.

\begin{table}[!htbp]
\centering
\caption{Qwen3-8B-Base throughput (tok/s) on H100. Parentheses give throughput relative to FA3 at the same length. Decode measurements are reported at extended lengths; dashes denote cells not included in this table. FA4 is compared for prefill only.}
\label{tab:efficiency}
\setlength{\tabcolsep}{3pt}
\renewcommand{\arraystretch}{1.05}
\small
\begin{tabular*}{\textwidth}{@{\extracolsep{\fill}} l rrrr @{}}
\toprule
\textbf{Method} & \multicolumn{1}{c}{4K} & \multicolumn{1}{c}{32K} & \multicolumn{1}{c}{64K} & \multicolumn{1}{c}{128K} \\
\midrule
\multicolumn{5}{l}{\textbf{Prefill}} \\
FA2 & 31211 ($0.92\times$) & 17796 ($0.78\times$) & 12238 ($0.69\times$) & 7433 ($0.61\times$) \\
FA3 & 33915 ($1.00\times$) & 22869 ($1.00\times$) & 17701 ($1.00\times$) & 12269 ($1.00\times$) \\
FA4 & 33455 ($0.99\times$) & 22358 ($0.98\times$) & 16690 ($0.94\times$) & 11400 ($0.93\times$) \\
CuTe (without bifocal attention) & 33235 ($0.98\times$) & 22431 ($0.98\times$) & 16860 ($0.95\times$) & 11494 ($0.94\times$) \\
Jet-Long (unfused) & 31242 ($0.92\times$) & 17810 ($0.78\times$) & 10909 ($0.62\times$) & 6932 ($0.57\times$) \\
Jet-Long CuTe & 33830 ($1.00\times$) & 22759 ($1.00\times$) & 15612 ($0.88\times$) & 10382 ($0.85\times$) \\
\midrule
\multicolumn{5}{l}{\textbf{Decode}} \\
FA2 & --- & --- & 74.80 ($1.09\times$) & 60.16 ($1.07\times$) \\
FA3 & --- & --- & 68.33 ($1.00\times$) & 56.05 ($1.00\times$) \\
CuTe (without bifocal attention) & --- & --- & 71.48 ($1.05\times$) & 58.38 ($1.04\times$) \\
Jet-Long (unfused) & --- & --- & 14.20 ($0.21\times$) & 8.32 ($0.15\times$) \\
Jet-Long CuTe & --- & --- & 73.90 ($1.08\times$) & 58.03 ($1.04\times$) \\
\bottomrule
\end{tabular*}
\end{table}

Across Qwen3-1.7B/4B/8B at 64K--128K, fused prefill reaches $0.83$--$0.88\times$ FA3 throughput, $0.88$--$0.93\times$ the matched CuTe control, and $1.43$--$1.56\times$ the unfused implementation. Thus, the bifocal attention rule incurs a $7$--$12\%$ prefill throughput cost relative to the matched control; the $1.28$--$1.45\times$ gain over FA2 largely reflects the underlying CuTe implementation. Fusion substantially reduces the cost of bifocal attention, while a measurable prefill cost remains relative to FA3.

For Qwen3-8B at 64K--128K, fused decode reaches $1.04$--$1.08\times$ FA3 and $0.994$--$1.034\times$ the matched control, while remaining within $3.6\%$ of FA2 throughput. At native lengths, $G=1$ bypasses extension; switching this fallback from FA2 to FA3 yields $0.97$--$1.00\times$ FA3 throughput across sizes. The extended-length branch is unchanged, and the measurements above use the same frozen stack throughout.

\section{Conclusion}
\label{sec:conclusion}
We presented \textbf{Jet-Long}, a tuning-free zero-shot context-extension method combining a protected local RoPE window, analytically determined remote grouping, and cache-aware fused execution. On Qwen3-1.7B/4B/8B-Base up to 128K, it improves RULER over the strongest compared baseline by $4.79$/$2.18$/$2.03$ pp, achieves the best aggregate HELMET-RAG score, and the lowest PG-19 perplexity at each size. Three-seed evaluations further support transfer to Meta-Llama-3-8B and post-trained Qwen3, while Jet-Nemotron demonstrates applicability to hybrid attention. Fusion reaches $0.83$--$0.88\times$ FA3 prefill throughput at extended lengths across the three Qwen3 sizes; a matched CuTe control isolates the remaining routing cost. Correction rotations avoid cache rewrites but retain historical hidden states, an approximation assessed in Appendix~\ref{app:cache_validation}.

Jet-Long addresses position-OOD at the RoPE level, while the complementary attention-diffusion failure mode is naturally tackled by architectural alternatives~\citep{beltagy2020longformer,zaheer2020bigbird,katharopoulos2020lineartransformer,choromanski2021performer,gu2023mamba}. Because Jet-Long requires a softmax-with-RoPE base, natural extensions include other softmax-with-RoPE variants such as Multi-head Latent Attention~\citep{liu2024deepseek} and sparse attention, and architectures interleaving softmax with sparse or linear-attention sub-layers, beyond the Jet-Nemotron experiments in Section~\ref{sec:hybrid}.

\begin{ack}
This work received no third-party funding. We declare no other competing interests.
\end{ack}

\clearpage
{
\small
\setlength{\bibsep}{1pt}

\bibliographystyle{plainnat}
\bibliography{ref}
}


\clearpage
\appendix

\section{Limitations}
\label{app:limitations}

\paragraph{Comparison scope.}
Our controlled comparisons focus on training-free extension. We do not provide a matched comparison with training-based systems such as LongRoPE~\citep{ding2024longrope}, nor a direct comparison with GALI~\citep{li2025training} or ReRoPE~\citep{su2023rerope}. Results reported for related baselines on other backbones and protocols cannot establish Jet-Long's relative performance against these methods. The Jet-Nemotron experiment establishes applicability to interleaved softmax and linear attention through comparison with the unextended models; it does not establish superiority over competing extension methods on hybrid architectures.

\paragraph{Evaluation coverage.}
RULER, HELMET-RAG, and PG-19 cover diagnostic long-context tasks, retrieval-augmented question answering, and language modeling. The post-trained experiments use chat templates but still evaluate RULER; they do not establish broad instruction-following, graph-reasoning, or agentic performance. The additional model families and evaluation through 1M expand the tested range, but do not guarantee transfer to arbitrary models, lengths, or workloads. Reported seed variation reflects dataset sampling rather than training variance, and mean $\pm$ half-range is not a confidence interval.

\paragraph{Cache-history approximation.}
Correction rotations recover the phases required by the current group size for the cached key vectors. They do not recompute historical hidden states or values after $G$ changes. Retained-cache decoding therefore differs from full re-prefill. Appendix~\ref{app:cache_validation} characterizes this difference at a tested boundary and in a stratified task analysis; these finite observations do not guarantee unchanged decisions at every transition.

\paragraph{Positional resolution and future adaptation.}
Discrete aliasing preserves separate token content but removes within-group positional distinctions from remote RoPE coordinates. Its aggregate advantage over frequency interpolation does not hold on every subtask, and the mapping comparison is limited to 64K and 128K. Distance-adaptive combinations of interpolation and aliasing, and long-context fine-tuning with Jet-Long, remain untested. Fine-tuning would also require evaluating short-context retention and generalization across group sizes.

\paragraph{Systems and extreme-length constraints.}
Jet-Long retains dense attention: fusion reduces implementation overhead but does not remove quadratic prefill computation or linear KV-cache storage. Throughput depends on hardware, kernels, and workload; our H100 comparisons include a measurable routing cost relative to a matched CuTe control. At 512K and 1M, the standard Self-Extend implementation could not be evaluated practically under our memory and runtime constraints (Section~\ref{sec:extreme}), so the extreme-length comparison covers a smaller set of methods.

\section{Metric definitions}
\label{app:metrics}

\paragraph{Percentage points (pp).}
For two scores $a_1, a_2 \in [0, 100]$ reported as percentages, the absolute gap in percentage points is
\begin{equation}
\Delta_{\text{pp}} = a_1 - a_2,
\end{equation}
which we use throughout the paper to describe RULER and HELMET-RAG accuracy differences. This is distinct from the relative percent change $(a_1 - a_2)/a_2$.

\paragraph{Perplexity (ppl).}
For a held-out token sequence $x_1, \dots, x_T$ scored under a language model with conditional probabilities $p(x_t \mid x_{<t})$, perplexity is the exponentiated mean negative log-likelihood per token,
\begin{equation}
\text{ppl} = \exp\!\left(-\frac{1}{T}\sum_{t=1}^{T} \log p(x_t \mid x_{<t})\right);
\end{equation}
lower is better. On PG-19 we use anchored growing-window evaluation: at each context length $L$ in $\{4\text{K}, 8\text{K}, 16\text{K}, 32\text{K}, 64\text{K}, 96\text{K}, 128\text{K}\}$, the next $1024$ tokens are scored conditional on the preceding $L$ tokens, and ppl is averaged across the 100 books in the PG-19 evaluation split.

\paragraph{Geometric-mean aggregation across lengths.}
For each model, PG-19 perplexity is aggregated by the geometric mean across lengths: $n=6$ for the main comparison (4K, 8K, 16K, 32K, 64K, 128K), and $n=7$ for the appendix table that additionally includes 96K,
\begin{equation}
\overline{\text{ppl}} = \left(\prod_{i=1}^{n} \text{ppl}_i\right)^{1/n} = \exp\!\left(\frac{1}{n}\sum_{i=1}^{n} \log \text{ppl}_i\right),
\end{equation}
This averages log perplexity uniformly across lengths. The Avg columns spanning model sizes in Table~\ref{tab:ruler_helmet} are arithmetic means of the three displayed model scores. The six-length main and seven-length appendix aggregates use different length sets and should not be equated.

\paragraph{Dataset-sampling variation.}
For the repeated RULER evaluations, let $a_s$ be the mean task score for dataset seed $s$. We report the mean across seeds and the half-range,
\begin{equation}
\bar a = \frac{1}{m}\sum_{s=1}^{m}a_s,\qquad
h = \frac{\max_s a_s-\min_s a_s}{2},\qquad m=3.
\end{equation}
The notation $\bar a\pm h$ describes variation across independently generated evaluation datasets. It is not a standard deviation, standard error, confidence interval, or estimate of training variance. The cross-family and post-trained comparisons use seeds 42/44/45; the extreme-length evaluation also uses three dataset seeds. The main Base-model tables and the additional single-seed sensitivity tests do not carry these error bars.

\section{Cross-family validation}
\label{app:cross_family}

We evaluate Meta-Llama-3-8B-Base, whose native context window is 8K and RoPE base is $\theta=500{,}000$, complementing the Qwen3 experiments. Table~\ref{tab:llama3_ruler} reports all seven lengths from 4K to 128K, with means and half-ranges over three independently generated RULER datasets (seeds 42, 44, and 45), using 500 examples for each of the 13 tasks per seed and length. The half-range is $(\max_s a_s-\min_s a_s)/2$, where $a_s$ is the aggregate score for dataset seed $s$. It measures dataset-sampling variation rather than training variance or a confidence interval; the same convention applies to the post-trained table below and the extreme-length results in Section~\ref{sec:extreme}.

Jet-Long's mean at 64K is 0.71~pp below Self-Extend, with overlapping reported spreads. At 96K and 128K, it exceeds the strongest compared baseline by 4.50 and 11.99~pp, respectively. Self-Extend with $G=19$ leads at 32K, while Jet-Long leads at 16K. Within the 8K native window, Jet-Long mathematically recovers base attention; its measured means differ from the unextended model by 0.21 and 0.16~pp at 4K and 8K. We report these implementation-level measurements without claiming numerically identical outputs.

\begin{table}[htbp]
\centering
\caption{RULER on Meta-Llama-3-8B-Base (native window 8K): mean $\pm$ half-range over three dataset seeds at all seven lengths. DCA uses chunk size $s$ and Self-Extend uses group size $G$; both use local window $w=1{,}024$. Bold marks the highest mean at each length.}
\label{tab:llama3_ruler}
\scriptsize
\setlength{\tabcolsep}{1.3pt}
\renewcommand{\arraystretch}{1.08}
\begin{tabular*}{\textwidth}{@{\extracolsep{\fill}} l ccccccc @{}}
\toprule
\textbf{Method} & \textbf{4K} & \textbf{8K} & \textbf{16K} & \textbf{32K} & \textbf{64K} & \textbf{96K} & \textbf{128K} \\
\midrule
Unextended & $\mathbf{93.36}\pm0.04$ & $\mathbf{91.54}\pm0.55$ & $0.03\pm0.02$ & $0.00\pm0.00$ & $0.00\pm0.00$ & $0.00\pm0.00$ & $0.00\pm0.00$ \\
DCA ($s=5{,}120$) & $93.26\pm0.17$ & $89.18\pm0.35$ & $69.45\pm0.81$ & $56.16\pm0.78$ & $30.30\pm0.40$ & $16.32\pm0.63$ & $12.29\pm0.25$ \\
DCA ($s=7{,}168$) & $93.25\pm0.18$ & $91.49\pm0.52$ & $65.43\pm0.45$ & $41.97\pm0.56$ & $13.58\pm0.59$ & $8.41\pm0.32$ & $6.00\pm0.18$ \\
DNTK & $93.26\pm0.19$ & $91.50\pm0.50$ & $84.31\pm0.36$ & $64.95\pm0.54$ & $28.94\pm0.64$ & $19.32\pm0.11$ & $11.59\pm0.28$ \\
Self-Extend ($G=32$) & $88.46\pm0.17$ & $84.47\pm0.55$ & $81.46\pm0.71$ & $73.38\pm0.22$ & $\mathbf{58.28}\pm0.97$ & $47.86\pm1.38$ & $31.86\pm0.99$ \\
Self-Extend ($G=19$) & $89.52\pm0.18$ & $86.28\pm0.50$ & $83.24\pm0.60$ & $\mathbf{74.90}\pm0.38$ & $53.99\pm0.64$ & $34.20\pm0.53$ & $14.64\pm0.20$ \\
YaRN & $77.14\pm0.57$ & $63.39\pm0.77$ & $57.34\pm0.13$ & $37.80\pm0.42$ & $31.87\pm0.37$ & $26.15\pm0.89$ & $12.21\pm0.79$ \\
Jet-Long & $93.15\pm0.22$ & $91.38\pm0.61$ & $\mathbf{86.96}\pm0.28$ & $73.18\pm0.93$ & $57.57\pm1.17$ & $\mathbf{52.36}\pm1.64$ & $\mathbf{43.85}\pm0.67$ \\
\bottomrule
\end{tabular*}
\end{table}

\section{Post-trained checkpoints}
\label{app:posttrained}

Table~\ref{tab:posttrained_ruler} evaluates the post-trained Qwen3-1.7B and Qwen3-8B checkpoints using their chat templates across all seven lengths from 4K to 128K. Each entry aggregates all 13 RULER tasks with 500 examples per task on each of three independently generated datasets (seeds 42, 44, and 45). These checkpoints and prompts differ from the Base-checkpoint experiments in the main comparison.

Jet-Long has the highest mean at 64K, 96K, and 128K on Qwen3-1.7B. On Qwen3-8B, DNTK leads at 64K, while Jet-Long has the highest mean at 96K and 128K. Within the native 32K window, Jet-Long's measured means differ from the unextended model by at most 0.17~pp for 1.7B and 0.11~pp for 8B, despite the attention rules being mathematically equivalent. Daggers identify baseline cells with at least 10\% empty generations in at least one seed; these scores reflect generation failures as well as task capability. DCA and Jet-Long have no such flags. The results demonstrate prompted long-context performance on RULER, with the broader evaluation limits discussed in Appendix~\ref{app:limitations}.

\begin{table}[htbp]
\centering
\caption{RULER on post-trained Qwen3 checkpoints with their chat templates: mean $\pm$ half-range over three dataset seeds at all seven lengths. Bold marks the highest mean for each model and length. $\dagger$ denotes an empty-generation rate of at least 10\% in at least one seed.}
\label{tab:posttrained_ruler}
\scriptsize
\setlength{\tabcolsep}{2pt}
\renewcommand{\arraystretch}{1.08}
\begin{tabular*}{\textwidth}{@{\extracolsep{\fill}} l ccccccc @{}}
\toprule
\textbf{Method} & \textbf{4K} & \textbf{8K} & \textbf{16K} & \textbf{32K} & \textbf{64K} & \textbf{96K} & \textbf{128K} \\
\midrule
\multicolumn{8}{l}{\textbf{Qwen3-1.7B (post-trained)}} \\
Unextended & $81.50\pm0.04$ & $77.56\pm0.73$ & $\mathbf{70.25}\pm1.28$ & $\mathbf{63.85}\pm0.87$ & $42.89\pm0.31$ & $13.15\pm0.42$ & $2.37\pm1.11$ \\
DCA & $\mathbf{81.51}\pm0.08$ & $77.60\pm0.84$ & $70.20\pm1.32$ & $57.85\pm0.49$ & $44.13\pm0.44$ & $41.07\pm0.65$ & $38.51\pm1.03$ \\
DNTK & $81.49\pm0.12$ & $\mathbf{77.63}\pm0.71$ & $70.21\pm1.40$ & $63.82\pm1.00$ & $51.78\pm0.35$ & $36.28\pm0.32^{\dagger}$ & $27.70\pm1.04^{\dagger}$ \\
Self-Extend & $80.86\pm0.10$ & $75.49\pm1.00$ & $69.11\pm1.95$ & $57.83\pm0.90$ & $48.51\pm1.76^{\dagger}$ & $48.30\pm0.49$ & $42.75\pm0.87^{\dagger}$ \\
YaRN & $72.06\pm1.59$ & $60.18\pm0.49$ & $17.21\pm1.35^{\dagger}$ & $27.45\pm1.63^{\dagger}$ & $27.26\pm0.38^{\dagger}$ & $34.36\pm0.14^{\dagger}$ & $25.78\pm0.19^{\dagger}$ \\
Jet-Long & $81.38\pm0.07$ & $77.48\pm0.76$ & $70.11\pm1.41$ & $63.68\pm0.81$ & $\mathbf{54.88}\pm1.23$ & $\mathbf{50.50}\pm0.89$ & $\mathbf{47.99}\pm0.90$ \\
\midrule
\multicolumn{8}{l}{\textbf{Qwen3-8B (post-trained)}} \\
Unextended & $84.60\pm0.18$ & $78.83\pm1.09$ & $81.21\pm1.33$ & $77.59\pm0.46$ & $61.22\pm0.41$ & $19.70\pm0.59$ & $3.00\pm0.20$ \\
DCA & $84.60\pm0.18$ & $78.84\pm1.15$ & $\mathbf{81.23}\pm1.50$ & $76.29\pm0.62$ & $64.57\pm0.43$ & $62.78\pm0.23$ & $61.05\pm0.46$ \\
DNTK & $84.63\pm0.18$ & $78.88\pm1.07$ & $81.21\pm1.39$ & $\mathbf{77.76}\pm0.32$ & $\mathbf{77.63}\pm0.73$ & $70.56\pm0.70$ & $63.62\pm0.64$ \\
Self-Extend & $82.36\pm0.12$ & $77.75\pm1.02$ & $77.75\pm0.42$ & $76.81\pm0.42$ & $73.65\pm0.79$ & $71.81\pm0.88$ & $68.54\pm0.16$ \\
YaRN & $\mathbf{84.81}\pm0.29$ & $\mathbf{81.03}\pm0.96$ & $74.17\pm0.50$ & $76.19\pm0.35$ & $64.85\pm0.46$ & $61.54\pm0.77$ & $56.92\pm0.68$ \\
Jet-Long & $84.51\pm0.17$ & $78.81\pm1.13$ & $81.19\pm1.47$ & $77.48\pm0.39$ & $74.91\pm0.23$ & $\mathbf{72.94}\pm0.36$ & $\mathbf{68.88}\pm0.21$ \\
\bottomrule
\end{tabular*}
\end{table}

\section{Cache correction and group-size transitions}
\label{app:cache_validation}

\paragraph{Retained cache versus full recomputation.}
The correction rotations in Section~\ref{sec:kv-cache} exactly reconstruct the current RoPE phases for stored key vectors. Historical hidden states and values remain unchanged, so this operation is not equivalent to recomputing the entire prefix under the new $G$. We compare retained-cache decoding (A) with complete re-prefill at every step (B) on Qwen3-1.7B-Base at the length boundary $L=98{,}304\rightarrow98{,}305$, where $w_{\mathrm{pretrained}}=32{,}768$ and $G$ changes from 3 to 4.

Table~\ref{tab:cache_reprefill} shows a larger logit discrepancy after the transition, while all tested greedy decisions agree. At the boundary itself, the maximum absolute logit difference increases from 0.1719 to 0.8750. The nonzero discrepancy in the no-transition control reflects numerical differences between prefill and decode kernels. These observations characterize a finite test; they do not establish equivalence of the two decoding procedures.

\begin{table}[htbp]
\centering
\caption{Retained-cache decoding (A) versus full re-prefill (B) around a $G:3\rightarrow4$ transition on Qwen3-1.7B-Base. Agreement counts refer to the tested greedy next-token decisions.}
\label{tab:cache_reprefill}
\small
\setlength{\tabcolsep}{7pt}
\begin{tabular}{lccc}
\toprule
\textbf{Regime} & \textbf{Mean $\max|\Delta\mathrm{logit}|$} & $\mathrm{KL}(A\parallel B)$ & \textbf{Argmax agreement} \\
\midrule
Before transition & 0.1781 & $\sim10^{-3}$ & 5/5 \\
After transition & 0.8000 & $1.84\times10^{-2}$ & 5/5 \\
No-transition control & 0.2094 & $\sim5\times10^{-4}$ & 10/10 \\
\bottomrule
\end{tabular}
\end{table}

\paragraph{Stratified task analysis.}
We also analyze a subset of RULER generations containing 9,247 examples that cross a $G$ boundary. To account for completion length, we stratify by model, task, target length, and completion-length bin, retaining 49 bins with at least 20 examples per arm. Table~\ref{tab:transition_scores} reports the crossed-minus-non-crossed score difference. Its mean is $-0.74$~pp and its reported 95\% confidence interval includes zero; the median is zero, and crossing scores lower in 24 of 49 bins. This observational comparison does not identify a causal transition effect or rule out degradation on other workloads.

\begin{table}[htbp]
\centering
\caption{Stratified RULER analysis of group-size transitions. Differences are scores for crossed minus non-crossed generations.}
\label{tab:transition_scores}
\small
\setlength{\tabcolsep}{10pt}
\begin{tabular}{lr}
\toprule
\textbf{Statistic} & \textbf{Result} \\
\midrule
Mean score difference & $-0.74$~pp \\
95\% confidence interval & $[-1.83,+0.35]$~pp \\
Median score difference & $0.00$~pp \\
Bins where crossing scored lower & 24/49 \\
\bottomrule
\end{tabular}
\end{table}

\section{Baseline sensitivity}
\label{app:baseline_sensitivity}

The configurations in Appendix~\ref{app:baseline_hp} support the main Qwen3 Base-model comparisons. Table~\ref{tab:baseline_sensitivity} provides a separate controlled rerun on Qwen3-1.7B-Base (500 examples per task, seed 42), testing additional DCA and Self-Extend configurations held fixed across 64K, 96K, and 128K. Here $s/w$ denotes DCA's chunk/local-window sizes and $G/w$ denotes Self-Extend's group/window sizes.

Reducing Self-Extend's group size from 8 to 5 improves its score by 3.07--3.93~pp; Jet-Long still leads the best compared configuration by 6.37, 3.53, and 5.81~pp at 64K, 96K, and 128K. Increasing DCA's chunk size does not improve this comparison. The Llama-family configurations in Table~\ref{tab:llama3_ruler} also illustrate that a setting's effect depends on the model: the alternative larger DCA chunk and smaller Self-Extend group both reduce the scores at 64K--128K. The smaller Self-Extend group instead improves the 16K and 32K scores. These tests assess sensitivity to the listed settings rather than an exhaustive hyperparameter search.

\begin{table}[htbp]
\centering
\caption{Additional baseline configurations on Qwen3-1.7B-Base, with 500 RULER examples per task and dataset seed 42. Each configuration is fixed across lengths.}
\label{tab:baseline_sensitivity}
\small
\setlength{\tabcolsep}{8pt}
\begin{tabular}{lccc}
\toprule
\textbf{Method and configuration} & \textbf{64K} & \textbf{96K} & \textbf{128K} \\
\midrule
DCA ($s=20{,}480$, $w=4{,}096$) & 48.80 & 43.45 & 42.62 \\
DCA ($s=28{,}672$, $w=4{,}096$) & 47.66 & 40.83 & 38.52 \\
Self-Extend ($G=8$, $w=1{,}024$) & 54.44 & 54.41 & 50.25 \\
Self-Extend ($G=5$, $w=1{,}024$) & 58.37 & 57.48 & 54.12 \\
Jet-Long & \textbf{64.74} & \textbf{61.01} & \textbf{59.93} \\
\bottomrule
\end{tabular}
\end{table}

\section{Additional lengths and detailed diagnostics}
\label{app:detailed_results}

The main Qwen3 comparisons use six lengths: 4K, 8K, 16K, 32K, 64K, and 128K. This appendix retains the additional 96K evaluations and the detailed task and window sweeps. Seven-length aggregates in this appendix include 96K and therefore differ from the six-length aggregates in the main text.

\subsection{PG-19 perplexity by length}
\label{app:ppl_by_length}

\begin{table}[t]
\centering
\caption{PG-19 perplexity by context length across three Qwen3 base models. Anchored growing-window ppl on 100 PG-19 books, stride 1024. Base shown in \textcolor{gray}{gray} as reference (collapses past its 32K native window). 
Best per column among extrapolation methods in \textbf{bold} (ties shared); the Avg column is the geometric mean across the seven lengths.}
\label{tab:ppl_by_length}
\setlength{\tabcolsep}{3pt}
\renewcommand{\arraystretch}{0.9}
\footnotesize
\begin{tabular}{l rrrrrrr c}
\toprule
\textbf{Method} & 4K & 8K & 16K & 32K & 64K & 96K & 128K & Avg \\
\midrule
\textbf{Qwen3-1.7B-Base} & \textcolor{gray}{11.52} & \textcolor{gray}{11.82} & \textcolor{gray}{11.59} & \textcolor{gray}{12.49} & \textcolor{gray}{12.60} & \textcolor{gray}{17.99} & \textcolor{gray}{71.00} & \multicolumn{1}{c}{\textcolor{gray}{16.39}} \\
+ DNTK            & \textbf{11.52} & \textbf{11.82} & \textbf{11.59} & \textbf{12.49} & 12.48 & 14.75 & 16.31 & \multicolumn{1}{c}{12.89} \\
+ YaRN            & 12.52 & 13.81 & 22.28 & 21.18 & 16.45 & 15.31 & 14.45 & \multicolumn{1}{c}{16.23} \\
+ DCA             & \textbf{11.52} & \textbf{11.82} & \textbf{11.59} & 12.50 & 11.65 & 11.78 & 11.59 & \multicolumn{1}{c}{11.77} \\
+ Self-Extend     & 11.55 & 11.91 & 11.74 & 12.69 & 11.74 & 11.83 & 11.51 & \multicolumn{1}{c}{11.85} \\
+ Jet-Long (ours) & \textbf{11.52} & \textbf{11.82} & \textbf{11.59} & \textbf{12.49} & \textbf{11.53} & \textbf{11.65} & \textbf{11.41} & \multicolumn{1}{c}{\textbf{11.71}} \\
\midrule
\textbf{Qwen3-4B-Base} & \textcolor{gray}{9.74} & \textcolor{gray}{9.90} & \textcolor{gray}{9.75} & \textcolor{gray}{10.45} & \textcolor{gray}{10.40} & \textcolor{gray}{21.44} & \textcolor{gray}{104.66} & \multicolumn{1}{c}{\textcolor{gray}{15.64}} \\
+ DNTK            & \textbf{9.74} & \textbf{9.90} & \textbf{9.75} & \textbf{10.45} & 11.43 & 12.88 & 13.93 & \multicolumn{1}{c}{11.05} \\
+ YaRN            & 10.12 & 11.10 & 11.11 & 14.37 & 13.33 & 13.48 & 12.98 & \multicolumn{1}{c}{12.27} \\
+ DCA             & \textbf{9.74} & \textbf{9.90} & \textbf{9.75} & \textbf{10.45} & 9.74 & 9.89 & 9.76 & \multicolumn{1}{c}{9.89} \\
+ Self-Extend     & 9.76 & 9.97 & 9.87 & 10.62 & 9.87 & 9.94 & 9.63 & \multicolumn{1}{c}{9.95} \\
+ Jet-Long (ours) & \textbf{9.74} & \textbf{9.90} & \textbf{9.75} & \textbf{10.45} & \textbf{9.67} & \textbf{9.81} & \textbf{9.62} & \multicolumn{1}{c}{\textbf{9.85}} \\
\midrule
\textbf{Qwen3-8B-Base} & \textcolor{gray}{8.64} & \textcolor{gray}{8.77} & \textcolor{gray}{8.61} & \textcolor{gray}{9.25} & \textcolor{gray}{9.18} & \textcolor{gray}{19.21} & \textcolor{gray}{79.37} & \multicolumn{1}{c}{\textcolor{gray}{13.56}} \\
+ DNTK            & \textbf{8.64} & \textbf{8.77} & \textbf{8.61} & \textbf{9.25} & 9.17 & 10.22 & 10.50 & \multicolumn{1}{c}{9.28} \\
+ YaRN            & 8.93 & 9.21 & 9.16 & 11.51 & 10.75 & 10.69 & 10.16 & \multicolumn{1}{c}{10.02} \\
+ DCA             & \textbf{8.64} & \textbf{8.77} & \textbf{8.61} & \textbf{9.25} & 8.70 & 8.81 & 8.68 & \multicolumn{1}{c}{8.78} \\
+ Self-Extend     & 8.66 & 8.83 & 8.70 & 9.38 & 8.77 & 8.80 & 8.55 & \multicolumn{1}{c}{8.81} \\
+ Jet-Long (ours) & \textbf{8.64} & \textbf{8.77} & \textbf{8.61} & \textbf{9.25} & \textbf{8.63} & \textbf{8.71} & \textbf{8.51} & \multicolumn{1}{c}{\textbf{8.73}} \\
\bottomrule
\end{tabular}
\end{table}

\subsection{RULER per-task diagnostics}
\label{app:ruler_tasks}

Table~\ref{tab:ruler_task_64K} gives the per-task diagnostic results at 64K. These diagnostics and the local-window sweep below are reported separately from the additional baseline-sensitivity runs in Appendix~\ref{app:baseline_sensitivity}.

\begin{table}[t]
\centering
\caption{RULER per-task accuracy at $L=65536$ on Qwen3-1.7B-Base and Qwen3-8B-Base. Tasks: S1/S2/S3 = Single-NIAH, MK1/MK2/MK3 = Multi-Key NIAH, MV/MQ = Multi-Value/Query NIAH, VT = Variable Tracking, CWE/FWE = Common/Frequent-Word Extraction, QA1/QA2 = Question Answering. Best per column in \textbf{bold} (ties shared).}
\label{tab:ruler_task_64K}
\setlength{\tabcolsep}{3pt}
\renewcommand{\arraystretch}{1.1}
\footnotesize
\begin{tabular}{l rrrrrrrrrrrrr r}
\toprule
\textbf{Method} & S1 & S2 & S3 & MK1 & MK2 & MK3 & MV & MQ & VT & CWE & FWE & QA1 & QA2 & \textbf{Avg} \\
\midrule
\multicolumn{15}{l}{\textbf{Qwen3-1.7B-Base}} \\
Base       & 99.80 & 49.60 & 82.40 & 47.20 & 10.40 & 0.20 & 62.10 & 58.15 & 48.56 & \textbf{0.58} & 52.00 & 25.60 & 22.20 & \multicolumn{1}{c}{42.98} \\
DNTK       & \textbf{100.00} & 93.00 & 99.40 & 78.40 & 16.40 & \textbf{1.60} & 81.60 & 81.05 & 54.24 & 0.38 & 70.93 & 41.20 & 26.80 & \multicolumn{1}{c}{57.31} \\
YaRN       & 98.40 & 75.40 & 89.00 & 67.60 & 6.00 & 0.00 & 49.65 & 46.80 & 0.00 & 0.56 & 23.13 & 21.20 & 23.20 & \multicolumn{1}{c}{38.53} \\
DCA        & \textbf{100.00} & 49.20 & 89.20 & 39.40 & 20.00 & 1.00 & 73.10 & 66.40 & 50.88 & 0.22 & \textbf{75.93} & \textbf{44.60} & 24.40 & \multicolumn{1}{c}{48.79} \\
Self-Extend  & \textbf{100.00} & 98.00 & 99.60 & 74.20 & 5.80 & 0.60 & \textbf{84.70} & \textbf{86.15} & 29.68 & 0.18 & 70.80 & 31.60 & 27.00 & \multicolumn{1}{c}{54.49} \\
Jet-Long   & \textbf{100.00} & \textbf{100.00} & \textbf{100.00} & \textbf{93.20} & \textbf{61.80} & \textbf{1.60} & 84.45 & 82.25 & \textbf{73.56} & 0.32 & 75.47 & 39.40 & \textbf{28.00} & \multicolumn{1}{c}{\textbf{64.62}} \\
\midrule
\multicolumn{15}{l}{\textbf{Qwen3-8B-Base}} \\
Base       & \textbf{100.00} & 92.20 & 79.00 & 79.40 & 41.60 & 12.40 & 83.15 & 72.15 & 92.80 & 28.30 & 86.13 & 41.20 & 28.60 & \multicolumn{1}{c}{64.38} \\
DNTK       & \textbf{100.00} & 99.80 & 99.60 & 91.80 & 78.80 & 28.60 & 93.10 & 93.90 & 98.88 & 16.84 & 81.67 & 54.80 & \textbf{41.80} & \multicolumn{1}{c}{75.35} \\
YaRN       & \textbf{100.00} & 97.80 & 96.20 & 70.00 & 45.20 & 7.20 & 89.85 & 79.00 & 94.16 & 27.62 & 77.07 & 55.20 & 38.60 & \multicolumn{1}{c}{67.53} \\
DCA        & \textbf{100.00} & 99.80 & 98.20 & 76.80 & 46.60 & 20.40 & 94.00 & 54.25 & 83.44 & 6.94 & \textbf{87.73} & 50.80 & 34.80 & \multicolumn{1}{c}{65.67} \\
Self-Extend  & \textbf{100.00} & 99.80 & 99.80 & 87.40 & 57.80 & 39.80 & 96.55 & \textbf{94.05} & \textbf{99.52} & 41.44 & 81.60 & \textbf{59.00} & 39.60 & \multicolumn{1}{c}{76.64} \\
Jet-Long   & \textbf{100.00} & \textbf{100.00} & \textbf{100.00} & \textbf{94.20} & \textbf{92.20} & \textbf{70.80} & \textbf{96.60} & 92.15 & 98.60 & \textbf{47.32} & 87.53 & 56.00 & 39.80 & \multicolumn{1}{c}{\textbf{82.71}} \\
\bottomrule
\end{tabular}
\end{table}

\subsection{Hybrid attention across seven lengths}
\label{app:hybrid_full}

Table~\ref{tab:ruler_jetlm_full} extends Table~\ref{tab:ruler_jetlm} with the 96K evaluation. Its averages include all seven lengths.

\begin{table}[t]
\centering
\caption{RULER accuracy by context length for Jet-Long applied to the hybrid Jet-Nemotron architecture~\citep{gu2025jet}, against the bare base model. The two configurations are mathematically identical within the 32K native window; the green tag reports the absolute accuracy-point gap of Jet-Long over Base for $L > 32$K.}
\label{tab:ruler_jetlm_full}
\setlength{\tabcolsep}{2.5pt}
\renewcommand{\arraystretch}{1.05}
\scriptsize
\begin{tabular}{l ccccccc c}
\toprule
\textbf{Method} & 4K & 8K & 16K & 32K & 64K & 96K & 128K & \emph{Avg} \\
\midrule
\textbf{Jet-Nemotron-2B} & 82.35 & 68.93 & 61.21 & 46.59 & 20.34 & 12.52 &  8.54 & \multicolumn{1}{c}{42.93} \\
+ Jet-Long & 82.35 & 68.93 & 61.21 & 46.58 & 40.33~\imp{19.99} & 37.40~\imp{24.88} & 33.78~\imp{25.24} & \multicolumn{1}{c}{52.94~\imp{10.01}} \\
\midrule
\textbf{Jet-Nemotron-4B} & 84.08 & 69.24 & 62.35 & 48.87 & 17.48 &  7.44 &  5.65 & \multicolumn{1}{c}{42.16} \\
+ Jet-Long & 84.08 & 69.24 & 62.35 & 48.88 & 39.93~\imp{22.45} & 36.69~\imp{29.25} & 33.14~\imp{27.49} & \multicolumn{1}{c}{53.47~\imp{11.31}} \\
\bottomrule
\end{tabular}
\end{table}

\subsection{Full local-window sweep}
\label{app:window_full}

Table~\ref{tab:w0_ablation_full} includes the additional 96K results and averages across the three extended lengths. The control $w_0=0$ retains self-attention in the local branch and leaves grouped remote attention active.

\begin{table}[t]
\centering
\caption{Full local-window ablation on Qwen3-4B/8B-Base at 64K, 96K, and 128K. Cells report RULER accuracy; \textcolor{red!60!black}{red} values give the gap in pp from the row's best score (\textbf{bold}). Avg aggregates the three extended lengths.}
\label{tab:w0_ablation_full}
\setlength{\tabcolsep}{1.2pt}
\renewcommand{\arraystretch}{0.9}
\scriptsize
\begin{tabular*}{\textwidth}{@{\extracolsep{\fill}} l ccccccc @{}}
\toprule
\textbf{Length} & $w_0{=}0$ & 256 & 512 & 1024 & 2048 & 4096 & 8192 \\
\midrule
\multicolumn{8}{l}{\textbf{Qwen3-4B-Base + Jet-Long}} \\
64K  & 4.10~\worse{-72.55} & \textbf{76.65} & 76.63~\worse{-0.02} & 76.49~\worse{-0.16} & 76.57~\worse{-0.09} & 75.91~\worse{-0.74} & 74.69~\worse{-1.96} \\
96K  & 2.07~\worse{-69.13} & 70.71~\worse{-0.48} & \textbf{71.19} & 70.77~\worse{-0.42} & 71.09~\worse{-0.11} & 70.57~\worse{-0.63} & 69.11~\worse{-2.09} \\
128K & 1.08~\worse{-64.55} & 64.13~\worse{-1.50} & \textbf{65.63} & 65.57~\worse{-0.05} & 64.74~\worse{-0.89} & 64.06~\worse{-1.56} & 63.52~\worse{-2.11} \\
\cmidrule(lr){1-8}
\emph{Avg} & 2.42~\worse{-68.74} & 70.50~\worse{-0.65} & \textbf{71.15} & 70.95~\worse{-0.21} & 70.80~\worse{-0.35} & 70.18~\worse{-0.97} & 69.11~\worse{-2.04} \\
\midrule
\multicolumn{8}{l}{\textbf{Qwen3-8B-Base + Jet-Long}} \\
64K  & 4.91~\worse{-78.60} & \textbf{83.51} & 83.23~\worse{-0.28} & 83.31~\worse{-0.20} & 83.24~\worse{-0.27} & 82.71~\worse{-0.80} & 81.40~\worse{-2.11} \\
96K  & 2.54~\worse{-72.96} & 74.29~\worse{-1.20} & 74.94~\worse{-0.56} & 75.20~\worse{-0.29} & \textbf{75.49} & 74.84~\worse{-0.65} & 73.82~\worse{-1.68} \\
128K & 1.55~\worse{-69.47} & 69.48~\worse{-1.55} & 70.49~\worse{-0.53} & 70.68~\worse{-0.35} & 71.01~\worse{-0.01} & \textbf{71.03} & 69.65~\worse{-1.38} \\
\cmidrule(lr){1-8}
\emph{Avg} & 3.00~\worse{-73.58} & 75.76~\worse{-0.82} & 76.22~\worse{-0.36} & 76.40~\worse{-0.19} & \textbf{76.58} & 76.19~\worse{-0.39} & 74.96~\worse{-1.63} \\
\bottomrule
\end{tabular*}
\end{table}

\section{Inference efficiency measurements and setup}
\label{app:efficiency}

Table~\ref{tab:efficiency_full} provides the reported Qwen3-8B-Base measurements from the revised H100 comparison, including the additional 96K length. These selected measurements complement Table~\ref{tab:efficiency}; the summary ranges in Section~\ref{sec:efficiency} cover Qwen3-1.7B/4B/8B. All ratios use FA3 as the reference. The matched CuTe control without bifocal attention omits Jet-Long's local/remote attention treatment while retaining the underlying backend. As explained in Section~\ref{sec:efficiency}, comparing it with fused Jet-Long measures the cost of the context-extension rule, while comparing fused and unfused Jet-Long at extended lengths measures the benefit of optimizing that rule.

\begin{table}[!htbp]
\centering
\caption{Selected Qwen3-8B-Base prefill and decode throughput (tok/s) on H100 from the revised comparison. Parentheses give throughput relative to FA3 at the matching length. Dashes denote unreported decode cells; FA4 is compared for prefill only.}
\label{tab:efficiency_full}
\setlength{\tabcolsep}{2pt}
\renewcommand{\arraystretch}{1.05}
\scriptsize
\begin{tabular*}{\textwidth}{@{\extracolsep{\fill}} l rrrrr @{}}
\toprule
\textbf{Method} & \multicolumn{1}{c}{4K} & \multicolumn{1}{c}{32K} & \multicolumn{1}{c}{64K} & \multicolumn{1}{c}{96K} & \multicolumn{1}{c}{128K} \\
\midrule
\multicolumn{6}{l}{\textbf{Prefill}} \\
FA2 & 31211 ($0.92\times$) & 17796 ($0.78\times$) & 12238 ($0.69\times$) & 9332 ($0.64\times$) & 7433 ($0.61\times$) \\
FA3 & 33915 ($1.00\times$) & 22869 ($1.00\times$) & 17701 ($1.00\times$) & 14615 ($1.00\times$) & 12269 ($1.00\times$) \\
FA4 & 33455 ($0.99\times$) & 22358 ($0.98\times$) & 16690 ($0.94\times$) & 13693 ($0.94\times$) & 11400 ($0.93\times$) \\
CuTe (without bifocal attention) & 33235 ($0.98\times$) & 22431 ($0.98\times$) & 16860 ($0.95\times$) & 13810 ($0.94\times$) & 11494 ($0.94\times$) \\
Jet-Long (unfused) & 31242 ($0.92\times$) & 17810 ($0.78\times$) & 10909 ($0.62\times$) & 8548 ($0.58\times$) & 6932 ($0.57\times$) \\
Jet-Long CuTe & 33830 ($1.00\times$) & 22759 ($1.00\times$) & 15612 ($0.88\times$) & 12575 ($0.86\times$) & 10382 ($0.85\times$) \\
\midrule
\multicolumn{6}{l}{\textbf{Decode}} \\
FA2 & --- & --- & 74.80 ($1.09\times$) & 67.01 ($1.09\times$) & 60.16 ($1.07\times$) \\
FA3 & --- & --- & 68.33 ($1.00\times$) & 61.44 ($1.00\times$) & 56.05 ($1.00\times$) \\
CuTe (without bifocal attention) & --- & --- & 71.48 ($1.05\times$) & 64.14 ($1.04\times$) & 58.38 ($1.04\times$) \\
Jet-Long (unfused) & --- & --- & 14.20 ($0.21\times$) & 10.49 ($0.17\times$) & 8.32 ($0.15\times$) \\
Jet-Long CuTe & --- & --- & 73.90 ($1.08\times$) & 65.00 ($1.06\times$) & 58.03 ($1.04\times$) \\
\bottomrule
\end{tabular*}
\end{table}

\paragraph{Hardware, precision, and software.}
Measurements use an NVIDIA H100 80GB HBM3 (SM90), driver 535.216.03, BF16 model weights and Q/K/V, and FP32 softmax and LogSumExp statistics. The software stack is PyTorch 2.9.1+cu128, CUDA 12.8, \texttt{nvcc} 12.6.85, \texttt{nvidia-cutlass-dsl} 4.6.0.dev0, \texttt{quack-kernels} 0.5.3, \texttt{cuda-python} 13.3.1, and \texttt{apache-tvm-ffi} 0.1.13rc1. The attention implementations use the same Dao-AILab/flash-attention repository, with the following versions and pinned commits:

\begin{center}
\small
\begin{tabular}{@{}ll@{}}
FA2 & \texttt{v2.8.3} \\
FA3 (\texttt{hopper/}) & \texttt{00756db9d921da0846453283ddfbeb7457abd09b} \\
FA4 & \texttt{fa4-v4.0.0.beta23}, commit \texttt{2409214} \\
CUTLASS & \texttt{7127592069c2fe01b041e174ba4345ef9b279671} \\
\end{tabular}
\end{center}

\paragraph{Interpreting the controls.}
FA3 is the primary H100 baseline, and FA4 is a complementary implementation with an SM90 path. At 64K--128K across the three Qwen3 sizes, fused Jet-Long prefill reaches $0.90$--$0.94\times$ FA4 in addition to the FA3 and matched-control comparisons in Section~\ref{sec:efficiency}. The CuTe control itself improves on FA2, so a comparison with FA2 alone would conflate the routing cost with a change of backend. For Qwen3-8B decode, Jet-Long's throughput is close to the matched control and exceeds FA3 at the reported extended lengths. These throughput comparisons are specific to the measured hardware, software stack, and model configurations.

\paragraph{Native-window fallback.}
For $L\leq32$K, $G=1$ recovers standard attention and bypasses the extension code. The revised implementation uses FA3 for this path instead of the earlier FA2 fallback. The $L>32$K branch is unchanged; all reported revised measurements were rerun under the frozen stack above. This implementation change affects efficiency measurements, while the attention definition and accuracy results are unchanged.

\section{Baseline configurations}
\label{app:baseline_hp}

The four zero-shot baseline configurations in the main Qwen3 Base-model comparison are held constant across Qwen3-1.7B/4B/8B and the six main evaluation lengths, as outlined in Table~\ref{tab:baseline_hp}. These three Base checkpoints have a pretrained context window of $32{,}768$ tokens and a RoPE base $\theta=10^6$. Additional configurations are evaluated separately in Appendix~\ref{app:baseline_sensitivity}; Llama-specific DCA and Self-Extend settings appear in Table~\ref{tab:llama3_ruler}.

\begin{table}[t]
\centering
\caption{Baseline hyperparameters used for the comparisons in Section~\ref{sec:setup}. Held constant across Qwen3-1.7B/4B/8B and across evaluation lengths.}
\label{tab:baseline_hp}
\setlength{\tabcolsep}{4pt}
\renewcommand{\arraystretch}{1.1}
\footnotesize
\begin{tabular}{l >{\raggedright\arraybackslash}p{0.72\textwidth}}
\toprule
\textbf{Method} & \textbf{Configuration} \\
\midrule
DNTK~\citep{emozilla2023dyntk} & HuggingFace \texttt{rope\_type=dynamic}, \texttt{factor}$=4.0$, \texttt{original\_max\_position\_embeddings}$=32{,}768$; the NTK base $\beta$ is computed at runtime from the scaling factor over RoPE base $\theta = 10^6$ \\
YaRN~\citep{peng2023yarn} & HuggingFace \texttt{rope\_type=yarn}, \texttt{factor}$=4.0$, \texttt{max\_position\_embeddings}$=131{,}072$, \texttt{original\_max\_position\_embeddings}$=32{,}768$ (target context $32\text{K} \times 4 = 128\text{K}$) \\
DCA~\citep{an2024training} & \texttt{chunk\_size}$=20{,}480$, \texttt{local\_window}$=4{,}096$ \\
Self-Extend~\citep{jin2024llm} & \texttt{group\_size}$=8$, \texttt{window\_size}$=1{,}024$, \texttt{scale\_base}$=-1$ \\
\bottomrule
\end{tabular}
\end{table}

\section{Compute resources}
\label{app:compute}

The original Qwen3 Base-model and Jet-Nemotron experiments use NVIDIA H100 80GB Tensor Core GPUs. A RULER evaluation sweep over the 13 tasks, 500 samples each, across seven context lengths including the appendix-only 96K evaluation takes roughly 40--96 H100 GPU-hours per (model, method) pair, with the spread driven by model size (Qwen3-1.7B/4B/8B) and the longer-context tasks dominating the budget. Each HELMET-RAG run takes roughly 8--16 H100 GPU-hours per (model, method) pair, as does each PG-19 long-form perplexity sweep. The added cross-family and post-trained evaluations use three dataset seeds with 500 examples per task, while the 256K--1M study uses 100 examples per task and three seeds. These are additional evaluations beyond the original sweeps. Hardware and software for the revised throughput measurements are specified in Appendix~\ref{app:efficiency}.

\section{Broader impacts}
\label{app:broader}

\paragraph{Positive impacts.}
Jet-Long extends existing checkpoints without additional training, which can lower the training cost of adapting models to long-document QA, retrieval-augmented generation, and other long-input applications. Preserving the native-window attention rule allows the same checkpoint to support both short and extended inputs. The fused implementation reduces the overhead of the bifocal construction, although long-context inference still requires substantial memory and computation. Benefits for coding agents, energy consumption, or substitution of larger models are potential applications rather than outcomes directly measured in this work.

\paragraph{Potential negative impacts.} The long-context capabilities Jet-Long unlocks inherit the standard risks of the underlying open-weight base model (e.g., scaled disinformation, longer surveillance or data-extraction prompts, amplification of pretraining bias over longer contexts). One context-extension-specific concern is that extending the input window enables agents to ingest longer streams of personal data within a single session, which can amplify privacy risks if intermediate KV caches are not zeroed; downstream deployers should treat KV caches with the same retention policies as raw input. Jet-Long does not introduce new risk axes beyond those of the base checkpoint it is applied to and provides no additional access to model weights or training data; mitigation should rely on the standard practices around responsible deployment of the underlying open-weight LLMs.

\section{Licenses for existing assets}
\label{app:licenses}

Our experiments rely entirely on publicly released pretrained language models and benchmark datasets, all of which are properly cited in the main text. The Qwen3 base models~\citep{yang2025qwen3} we evaluate on (Qwen3-1.7B-Base, Qwen3-4B-Base, Qwen3-8B-Base) are released by Alibaba under the Apache License 2.0 (\url{https://huggingface.co/Qwen/Qwen3-8B-Base}), which permits research and commercial use with attribution. The hybrid Jet-Nemotron-2B and Jet-Nemotron-4B models~\citep{gu2025jet} are released by NVIDIA at \url{https://github.com/NVlabs/Jet-Nemotron}: the inference code is licensed under the Apache License 2.0, while the model weights are licensed under the NVIDIA non-commercial research license (\url{https://github.com/NVlabs/Jet-Nemotron/blob/main/LICENSE/jet_nemotron_models}), which permits research and evaluation but reserves commercial use to NVIDIA; our use is strictly for academic research and does not constitute commercial deployment, and we do not redistribute model weights. For benchmarks, RULER~\citep{hsieh2024ruler} is released by NVIDIA under the Apache License 2.0 (\url{https://github.com/NVIDIA/RULER}); HELMET~\citep{yen2024helmet} is released by Princeton NLP under the MIT License (\url{https://github.com/princeton-nlp/HELMET}); and PG-19~\citep{rae2019compressive} is released by Google DeepMind under the Apache License 2.0 (\url{https://github.com/google-deepmind/pg19}). All three benchmarks are used strictly for academic evaluation as intended by their creators. We do not redistribute any of these datasets and refer readers to the original sources for access.

The additional post-trained Qwen3 checkpoints and Qwen3-30B-A3B-Instruct-2507 are released under Apache License 2.0 (\url{https://huggingface.co/Qwen/Qwen3-8B}; \url{https://huggingface.co/Qwen/Qwen3-30B-A3B-Instruct-2507}). Meta-Llama-3-8B is distributed under the Meta Llama 3 Community License Agreement and its accompanying acceptable-use policy (\url{https://huggingface.co/meta-llama/Meta-Llama-3-8B}). We use the released checkpoints for evaluation and do not redistribute their weights.


\newpage
\section*{NeurIPS Paper Checklist}

\begin{enumerate}

\item {\bf Claims}
    \item[] Question: Do the main claims made in the abstract and introduction accurately reflect the paper's contributions and scope?
    \item[] Answer: \answerYes{}
    \item[] Justification: We elucidate our contributions in the Abstract and Introduction.

\item {\bf Limitations}
    \item[] Question: Does the paper discuss the limitations of the work performed by the authors?
    \item[] Answer: \answerYes{}
    \item[] Justification: Appendix~\ref{app:limitations}, the first appendix section, discusses comparison scope, evaluation coverage, the cache-history approximation, positional resolution, and systems constraints.

\item {\bf Theory assumptions and proofs}
    \item[] Question: For each theoretical result, does the paper provide the full set of assumptions and a complete (and correct) proof?
    \item[] Answer: \answerNA{}
    \item[] Justification: We do not have theoretical results.

    \item {\bf Experimental result reproducibility}
    \item[] Question: Does the paper fully disclose all the information needed to reproduce the main experimental results of the paper to the extent that it affects the main claims and/or conclusions of the paper (regardless of whether the code and data are provided or not)?
    \item[] Answer: \answerYes{}
    \item[] Justification: Experimental details are in Section~\ref{sec:setup}; baseline configurations and sensitivity tests are in Appendices~\ref{app:baseline_hp} and~\ref{app:baseline_sensitivity}. Section~\ref{sec:extreme} and Appendices~\ref{app:cross_family} and~\ref{app:posttrained} describe the additional evaluations, and Appendix~\ref{app:efficiency} gives the pinned efficiency stack.

\item {\bf Open access to data and code}
    \item[] Question: Does the paper provide open access to the data and code, with sufficient instructions to faithfully reproduce the main experimental results, as described in supplemental material?
    \item[] Answer: \answerYes{}
    \item[] Justification: The evaluated datasets are publicly available. The paper links the Jet-Long code repository, and the supplementary material includes a PyTorch implementation. Experimental settings and the pinned efficiency stack are described in the main text and appendix.

\item {\bf Experimental setting/details}
    \item[] Question: Does the paper specify all the training and test details (e.g., data splits, hyperparameters, how they were chosen, type of optimizer) necessary to understand the results?
    \item[] Answer: \answerYes{}
    \item[] Justification: This is discussed in Section~\ref{sec:setup}.

\item {\bf Experiment statistical significance}
    \item[] Question: Does the paper report error bars suitably and correctly defined or other appropriate information about the statistical significance of the experiments?
    \item[] Answer: \answerYes{}
    \item[] Justification: The cross-family, post-trained, and 256K--1M RULER experiments report mean and half-range over three independently generated datasets (Sections~\ref{sec:generalization} and~\ref{sec:extreme}; Appendices~\ref{app:cross_family} and~\ref{app:posttrained}). Appendix~\ref{app:metrics} defines this descriptive measure and distinguishes it from a confidence interval or training variance. The original Base-model tables and single-seed sensitivity tests do not report uncertainty; we do not claim statistical significance from their small margins.

\item {\bf Experiments compute resources}
    \item[] Question: For each experiment, does the paper provide sufficient information on the computer resources (type of compute workers, memory, time of execution) needed to reproduce the experiments?
    \item[] Answer: \answerYes{}
    \item[] Justification: Appendix~\ref{app:compute} reports H100 GPU-hour estimates for the original benchmark sweeps and sample counts for the added evaluations. Appendix~\ref{app:efficiency} specifies hardware, precision, software versions, and kernel commits for the revised throughput measurements.
    
\item {\bf Code of ethics}
    \item[] Question: Does the research conducted in the paper conform, in every respect, with the NeurIPS Code of Ethics \url{https://neurips.cc/public/EthicsGuidelines}?
    \item[] Answer: \answerYes{}
    \item[] Justification: We have carefully reviewed our research, ensuring it conforms with the NeurIPS Code of Ethics.

\item {\bf Broader impacts}
    \item[] Question: Does the paper discuss both potential positive societal impacts and negative societal impacts of the work performed?
    \item[] Answer: \answerYes{}
    \item[] Justification: Appendix~\ref{app:broader} discusses potential deployment benefits and negative impacts, distinguishing measured long-context results from possible benefits for agentic workflows or energy consumption.
    
\item {\bf Safeguards}
    \item[] Question: Does the paper describe safeguards that have been put in place for responsible release of data or models that have a high risk for misuse (e.g., pre-trained language models, image generators, or scraped datasets)?
    \item[] Answer: \answerNA{}
    \item[] Justification: At present, our proposed approach poses no immediate risk of misuse.

\item {\bf Licenses for existing assets}
    \item[] Question: Are the creators or original owners of assets (e.g., code, data, models), used in the paper, properly credited and are the license and terms of use explicitly mentioned and properly respected?
    \item[] Answer: \answerYes{}
    \item[] Justification: All pre-trained models and datasets used in our experiments are publicly available and properly cited in the main text. We discuss their licenses and our usage in Appendix~\ref{app:licenses}.

\item {\bf New assets}
    \item[] Question: Are new assets introduced in the paper well documented and is the documentation provided alongside the assets?
    \item[] Answer: \answerNA{}
    \item[] Justification: We do not release new datasets or model checkpoints. The accompanying implementation and linked repository provide a reproducibility artifact for the method; the experimental settings and dependencies are documented in the paper and appendix.

\item {\bf Crowdsourcing and research with human subjects}
    \item[] Question: For crowdsourcing experiments and research with human subjects, does the paper include the full text of instructions given to participants and screenshots, if applicable, as well as details about compensation (if any)? 
    \item[] Answer: \answerNA{}
    \item[] Justification: No crowdsourcing experiments and no research with human subjects.

\item {\bf Institutional review board (IRB) approvals or equivalent for research with human subjects}
    \item[] Question: Does the paper describe potential risks incurred by study participants, whether such risks were disclosed to the subjects, and whether Institutional Review Board (IRB) approvals (or an equivalent approval/review based on the requirements of your country or institution) were obtained?
    \item[] Answer: \answerNA{}
    \item[] Justification: No crowdsourcing experiments and no research with human subjects.

\item {\bf Declaration of LLM usage}
    \item[] Question: Does the paper describe the usage of LLMs if it is an important, original, or non-standard component of the core methods in this research? Note that if the LLM is used only for writing, editing, or formatting purposes and does \emph{not} impact the core methodology, scientific rigor, or originality of the research, declaration is not required.
    \item[] Answer: \answerNA{}
    \item[] Justification: LLMs were only used for grammar and for occasional code debugging.

\end{enumerate}

\end{document}